\documentclass[10pt,journal,compsoc]{IEEEtran}

\usepackage{xcolor,soul,framed} 

\colorlet{shadecolor}{yellow}
\usepackage[pdftex]{graphicx}
\graphicspath{{../pdf/}{../jpeg/}}
\DeclareGraphicsExtensions{.pdf,.jpeg,.png}

\usepackage[cmex10]{amsmath}
\usepackage{array}
\usepackage{mdwmath}
\usepackage{mdwtab}
\usepackage{eqparbox}
\usepackage{url}
\usepackage{graphicx}
\usepackage{amsmath}
\usepackage{amsfonts}
\usepackage{algorithm}
\usepackage{algpseudocode}
\usepackage{multirow}
\usepackage{footnote}
\usepackage{bm}
\usepackage{makecell}
\usepackage{pdflscape}
\usepackage{afterpage}
\usepackage{bbding}
\usepackage{lscape}
\usepackage{booktabs}
\usepackage{tikz}
\usepackage{threeparttable}
\usepackage{rotating, graphicx}

\usepackage{pgfplots}
\pgfplotsset{compat=1.12}
\usepackage{subcaption}
    \pgfplotsset{
        layers/my layer set/.define layer set={
            background,
            main,
            foreground
        }{ },
        set layers=my layer set,
    }

\usepackage[square,sort,comma,numbers]{natbib} 
\usepackage[top=0.4in,bottom=0.4in,left=0.4in,textwidth=7.7in]{geometry}

\usepackage[pagebackref=false,breaklinks=true,linkcolor=red,anchorcolor=blue, citecolor=green,letterpaper=true,colorlinks,bookmarks=false]{hyperref}
\newcommand{\etal}{\textit{et al}.}
\newcommand{\etc}{\textit{etc}}
\newcommand{\ie}{\textit{i}.\textit{e}.}
\newcommand{\eg}{\textit{e}.\textit{g}.}
\usepackage{amsfonts}

\usepackage{pifont}

\begin{document}

\title{Deep Learning for Visual Speech Analysis: A Survey}

\author{\IEEEauthorblockN{Changchong Sheng,
Gangyao Kuang*,
Liang Bai,
Chenping Hou,
Yulan Guo,
Xin Xu*, 
Matti Pietik{\"a}inen, and
Li Liu$^{*}$
\thanks{C. Sheng is with the College of Electronic Science and Technology, National University of Defense Technology (NUDT), China, and also with the National Key Laboratory of Electromagnetic Energy, Naval University of Engineering, China (e-mail: shengcc.nudt@gmail.com).
L. Liu, G. Kuang, and Y. Guo are with the College of Electronic Science and Technology, National University of Defense Technology (NUDT), China. 
L. Bai is with the College of Systems Engineering, NUDT, China. 
C. Hou is with the College of Liberal Arts and Sciences, NUDT, China. 
X. Xu is with the College of Intelligence Science and Technology, NUDT, China. 
M. Pietik{\"a}inen is with the Center for Machine Vision and Signal Analysis, Oulu University, Finland.}
\thanks{Corresponding authors: Li Liu (lilyliu\_nudt@163.com) and Gangyao Kuang (kuanggangyao@nudt.edu.cn) and Xin Xu (xinxu@nudt.edu.cn).}
\thanks{This work was partially supported by National Key R\&D Program of China No.2021YFB3100800, the Academy of Finland under grant 331883, the National Natural Science Foundation of China under Grant 62376283 and 61872379, and the Hubei Province Natural Science Foundation under grants 2023AFB446.}
}}

\markboth{IEEE TRANSACTIONS ON PATTERN ANALYSIS AND MACHINE INTELLIGENCE}
{Sheng \MakeLowercase{\textit{et al.}}: }

\IEEEtitleabstractindextext{%
\begin{abstract}
Visual speech, referring to the visual domain of speech, has attracted increasing attention due to its wide applications, such as public security, medical treatment, military defense, and film entertainment. As a powerful AI strategy, deep learning techniques have extensively promoted the development of visual speech learning. Over the past five years, numerous deep learning based methods have been proposed to address various problems in this area, especially automatic visual speech recognition and generation. To push forward future research on visual speech, this paper will present a comprehensive review of recent progress in deep learning methods on visual speech analysis. We cover different aspects of visual speech, including fundamental problems, challenges, benchmark datasets, a taxonomy of existing methods, and state-of-the-art performance. Besides, we also identify gaps in current research and discuss inspiring future research directions.
\end{abstract}

\begin{IEEEkeywords}
  Deep Learning, Visual Speech, Lip Reading, Speech Perception, Computer Vision, Computer Graphics
\end{IEEEkeywords}}

\maketitle
\IEEEdisplaynontitleabstractindextext
\IEEEpeerreviewmaketitle

\IEEEraisesectionheading{
\section{Introduction}
\label{sec:intro}}

\IEEEPARstart{H}{uman} speech is by nature bimodal: visual and audio. Visual speech refers to the visual domain of speech,~\ie, the movements of the lips, tongue, teeth, jaw,~\etc., and other facial muscles of the lower face that are naturally produced during talking~\cite{chen2001audiovisual}, while audio speech refers to the acoustic waveform pronounced by the speaker. Speech perception is intrinsically bimodal, as shown several decades ago by the famous McGurk effect~\cite{mcgurk1976hearing} that human speech perception depends not only on auditory information, but also on visual cues like lip movements. Therefore, undoubtedly, visual speech contributes to human speech perception, especially for people who are hearing-impaired or hard of hearing or when acoustic information is corrupted.

As a fundamental and challenging topic in computer vision and multimedia, automatic Visual Speech Analysis (VSA) has received increasing attention in recent years, due to the important role it plays in a wide variety of applications, many of which are newly emerging. 
VSA embraces two fundamental closely-related and formally-dual problems: Visual Speech Recognition (VSR) or Lip Reading, and Visual Speech Generation (VSG) or Lip Sequence Generation.
Significant progress has been witnessed in this field due to the recent boom of deep learning.
Typical academia and practical applications of VSA include multimodal speech recognition and enhancement~\cite{gabbay2017visual},  audio-to-video alignment~\cite{halperin2019dynamic}, audio speech synthesization~\cite{ephrat2017vid2speech, ephrat2017improved}, speaker recognition and verification~\cite{gabbay2017seeing,jia2018transfer}, medical assistance, security, forensic, video compression, entertainment, human-computer interaction, emotion understanding~\cite{ji2021audio,karras2017audio},~\etc.

To give some application examples, in speech recognition and enhancement, visual speech can be treated as a complementary signal to increase the accuracy and robustness of current audio speech recognition and separation under various unfavorable acoustic conditions~\cite{afouras2018deep,dupont2000audio}. 
In the medical domain, solving the VSR task can also help the hearing impaired~\cite{tye2007audiovisual} and people with vocal cord lesions. In public security, VSA can be applied to face forgery detection~\cite{haliassos2021lips} and liveness detection~\cite{akhtar2015biometric}. 
In human-computer interaction, visual speech can serve as a new type of interactive information, improving the diversity and robustness of interactions~\cite{sun2018lip}. In the entertainment domain, VSG technology plays a crucial role in personalized 3D talking avatars generation~\cite{zhou2020makelttalk} in virtual gaming and realizing high-fidelity photo-realistic talking videos generation for movie post-production like visual dubbing~\cite{garrido2015vdub}. In addition, VSR can be used to transcribe archival silent films.

\begin{figure*}[t]
    \centering
    \includegraphics[width=1.0\textwidth]{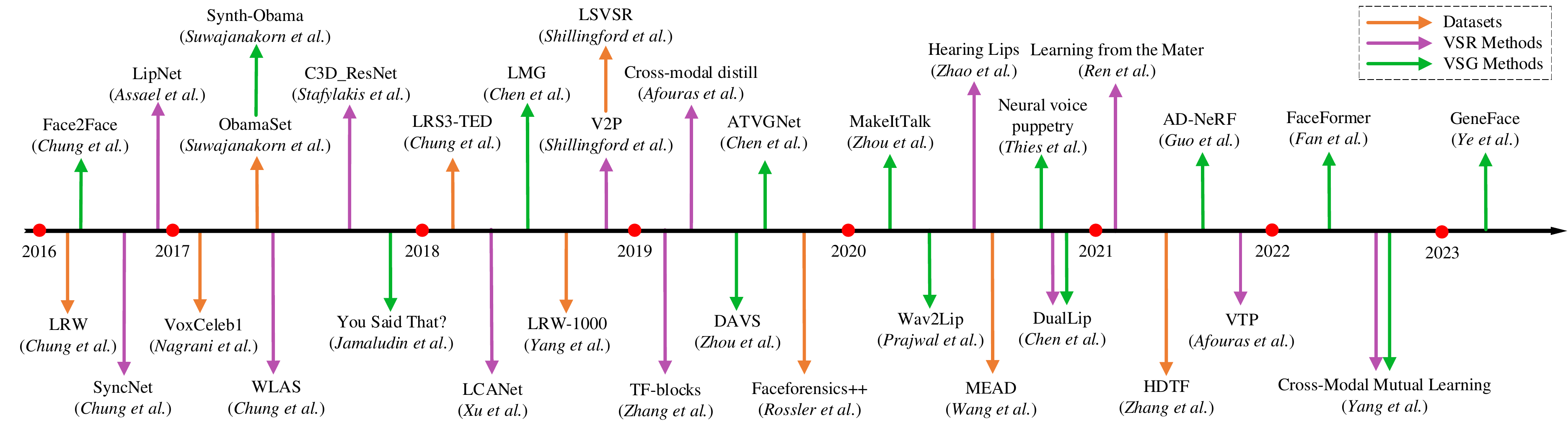} 
    \caption{Chronological milestones on visual speech analysis from 2016 to the present, including representative VSR and VSG methods, and audio-visual datasets. Handcrafted feature engineering methods dominated VSA until a transition took place in 2016 with the introduction of related deep networks.}
    \label{milestone}
    \end{figure*}

The core of VSA lies in visual speech representation learning and sequence modeling. In the era dominated by traditional VSA methods, shallow representations of visual speech such as visemes~\cite{potamianos2003recent}, mouth geometry descriptors~\cite{kirchho1999robust}, linear transformation features~\cite{potamianos1998image}, statistical representations~\cite{matthews2002extraction}, and sequence modeling like Gaussian process dynamical models~\cite{deena2010visual}, hidden Markov models (HMMs)~\cite{anderson2013expressive}, decision tree models~\cite{kim2015decision} were widely used in solving VSA tasks.   
Since the significant breakthroughs~\cite{AlexNet2012} of deep neural networks (DNNs) in the image classification task, most computer vision and natural language problems have focused explicitly on deep learning methods, including VSA. In 2016, deep learning based VSA methods~\cite{fan2015photo,chung2016lip} have vastly outperformed traditional approaches, bringing the VSA into the deep learning era. Meanwhile, the emergence of large-scale VSA datasets~\cite{chung2016lip,chung2017lip,chung2018VoxCeleb2,nagrani2017VoxCeleb,yang2019lrw} promoted the further development of deep learning based VSA research. In this paper, we mainly focus on the deep learning based VSA approaches. The milestones of VSA technologies from 2016 to the present are shown in Fig.~\ref{milestone}, including representative deep VSR and VSG methods and related audio-visual datasets.

Although the recent promising progress brought by deep learning in the past several years, the VSA technology is still in its early stages and incapable of performing at a level sufficient for real-world applications. 
This is certainly not because of a lack of effort by researchers, as there have been many excellent works on VSA~\cite{afouras2018deep,chen2020duallip,chung2017lip,ren2021learning,yang2022cross,zhao2020hearing}. Therefore, it is of great importance to systematically review the recent developments in the field, identify the main challenges and open problems preventing its advancement, and define promising future directions. However, a large part of VSA research remains rather scattered, and no such systematic surveys exist. This motivates this survey, which will fill this gap.

\subsection{The Scope of this Survey}

The main objective of this survey is to provide a comprehensive overview of current deep learning based VSA methods, in particular, VSR and VSG and related applications, main challenges, benchmark datasets, methods, and state-of-the-art (SOTA) results, together with existing gaps and promising future research directions.

There are three main reasons that we comprehensively overview VSR and VSG together. 
First, as the most fundamental problems in VSA, VSR and VSG cover most aspects of visual speech analysis. 
Other VSA-related tasks, such as speech enhancement, speaker verification, face forgery detection,~\etc., can be seen as extended applications of VSR and VSG. 
Second, because VSR and VSG are formal-dual and mutually promoted, the dual learning~\cite{he2016dual} and generative adversarial learning mechanisms~\cite{goodfellow2014generative} are widely adopted in many existing VSA works~\cite{chen2020duallip,prajwal2020lip,song2019talking,vougioukas2018end,wang2021one}. 
Therefore, we intend to provide a side-by-side perspective for readers to know the evolution of VSR and VSG. 
Third, VSR and VSG have common core technical points, such as visual speech representation learning approaches and contextual sequence modeling approaches. We hope it would be helpful for readers to have an accessible understanding of the cross-task transferability of these methods.

\subsection{Differences with Related Surveys}
Several surveys on VSA~\cite{chen2020comprises,fenghour2021deep,fernandez2018survey,mattheyses2015audiovisual,zhou2014review} have been published. However, they have only partially reviewed specific VSA tasks.
For example,~\cite{fenghour2021deep,fernandez2018survey,zhou2014review} conducted reviews on VSR, and~\cite{chen2020comprises,mattheyses2015audiovisual} focused on VSG. We give a brief conclusion related surveys and then emphasize our new contributions. 

In 2014, Zhou~\etal~\cite{zhou2014review} summarized three central questions of visual feature extraction for VSR: speaker dependency, head pose variation and efficient encoding of spatiotemporal information. Then, they reviewed mainstream visual features extraction and dynamic audio-visual speech fusion methods of VSR from the view of question-solving, which brought a new perspective for readers to know the developments of VSR. In 2015, Mattheyses~\etal~\cite{mattheyses2015audiovisual} gave an extensive and comprehensive overview of audio-visual speech synthesis with great effort. We advocate readers to refer to~\cite{mattheyses2015audiovisual} for the development history of VSG before 2015.
As~\cite{mattheyses2015audiovisual,zhou2014review} provided comprehensive surveys on traditional VSR and VSG methods, in this paper, we mainly focus on the recent advances driven by deep learning technologies. 
Adriana~\etal~\cite{fernandez2018survey} summarized VSR datasets according to the differences in recognition tasks and reviewed traditional and deep learning based methods of VSR. They mainly focused on the existing datasets and the analysis of VSR methods for different recognition tasks on each dataset. However, the research reviewed in~\cite{fernandez2018survey} is mainly pre-2018, prior to the recent striking success. Recently, Fenghour~\etal~\cite{fenghour2021deep} conducted a survey reviewing deep learning driven VSR methods, including audio-visual datasets, feature extraction, classification networks and classification schemes. However, some essential advances of VSR were omitted, such as self-supervised learning methods~\cite{arandjelovic2018objects,chung2019perfect,chung2020seeing,sheng2021cross}, cross-modal knowledge distillation methods~\cite{afouras2020asr,ma2021towards,yang2022cross}, graph neural networks backbone architectures~\cite{liu2020lip,sheng2021adaptive},~\etc. Chen~\etal~\cite{chen2020comprises} conducted a thoughtful analysis across several representative identity-independent VSG methods and designed a performance evaluation benchmark for VSG. However, their core contributions are well-defined standards of evaluation metrics rather than the comprehensive discussion and overview of VSG methods.

Now, we are in a place to summarize our main contributions in this paper.
\begin{itemize}
\renewcommand{\labelitemi}{$\bullet$}
    \item To the best of our knowledge, this is the \textbf{\emph{first}} survey paper to systematically and comprehensively review deep learning methods for visual speech analysis, covering two fundamental problems,~\ie, visual speech recognition and visual speech generation.
    \item Problem definition, main challenges, benchmark datasets and testing protocols are summarized for each problem, and notably, the relationship among different VSA problems is also identified.
    \item We propose a taxonomy to group the prominent methods. In addition, performance comparisons, merits, and demerits of representative approaches and their underlying connections are also analyzed.
    
    \item Open issues and promising directions in this field are provided.
\end{itemize}

\begin{figure}[h]
\centering
\footnotesize
\resizebox*{0.5\textwidth}{!}{\begin{tikzpicture}[xscale=0.8, yscale=0.36]

\draw [thick, -] (0, 16.5) -- (0, -1); \node [right] at (-0.5, 17) {\textbf{Deep Learning on Visual Speech}};
\draw [thick, -] (0, 16) -- (0.5, 16);\node [right] at (0.5, 16) {\textbf{Visual Speech Recognition~(Section~\ref{sec VSR})}};
\draw [thick, -] (1, 15.5) -- (1, 6);
\draw [thick, -] (1, 15) -- (1.5, 15);\node [right] at (1.5, 15) {Backbone architectures (Section~\ref{subsec Backbone})};
\draw [thick, -] (2, 14.5) -- (2,10);
\draw [thick, -] (2, 14) -- (2.5, 14);\node [right] at (2.5, 14) {Visual frontend network (Section~\ref{subsub vfn})};
\draw [thick, -] (3, 13.5) -- (3,11);
\draw [thick, -] (3, 13) -- (3.5, 13);\node [right] at (3.5, 13) {CNN based: VGG~\cite{chung2016lip}, ResNet~\cite{stafylakis2017combining}, STCNN~\cite{assael2016lipnet}...};
\draw [thick, -] (3, 12) -- (3.5, 12);\node [right] at (3.5, 12) {GCN based: STGCN~\cite{liu2020lip},ASST-GCN~\cite{sheng2021adaptive}};
\draw [thick, -] (3, 11) -- (3.5, 11);\node [right] at (3.5, 11) {Transformer based: VTP~\cite{prajwal2022sub}};

\draw [thick, -] (2, 10) -- (2.5, 10);\node [right] at (2.5, 10) {Temporal backend network (Section~\ref{subsub tbn})};

\draw [thick, -] (3, 9.5) -- (3, 7);
\draw [thick, -] (3, 9) -- (3.5, 9);\node [right] at (3.5, 9) {RNN based: BiLSTM, BiGRU, BiConvLSTM~\cite{wang2019multi}};
\draw [thick, -] (3, 8) -- (3.5, 8);\node [right] at (3.5, 8) {TCN based: MST-TCN~\cite{martinez2020lipreading}, DS-TCN~\cite{afouras2018deep_a}};
\draw [thick, -] (3, 7) -- (3.5, 7);\node [right] at (3.5, 7) {Transformer based: Transformer~\cite{afouras2018deep}, TF-blocks~\cite{zhang2019spatio}...}; 

\draw [thick, -] (1, 6) -- (1.5, 6);\node [right] at (1.5, 6) {Learning Paradigms~(Section~\ref{learning paradigms})};
\draw [thick, -] (2, 5.5) -- (2, 1);
\draw [thick, -] (2, 5) -- (2.5, 5);\node [right] at (2.5, 5) {Supervised learning (Section~\ref{sl for vsr})};

\draw [thick, -] (3, 4.5) -- (3, 2);
\draw [thick, -] (3, 4) -- (3.5, 4);\node [right] at (3.5, 4) {CTC based: TM-CTC~\cite{afouras2018deep}, LCANet~\cite{xu2018lcanet}};
\draw [thick, -] (3, 3) -- (3.5, 3);\node [right] at (3.5, 3) {seq2seq based: WAS~\cite{chung2017lip}, TM-seq2seq~\cite{afouras2018deep}};
\draw [thick, -] (3, 2) -- (3.5, 2);\node [right] at (3.5, 2) {KD based: self-KD~\cite{ma2021towards}, cross-modal KD~\cite{li2019improving}};

\draw [thick, -] (2, 1) -- (2.5, 1);\node [right] at (2.5, 1) {Self-supervised learning (Section~\ref{ul for vsr})};

\draw [thick, -] (3, 0.5) -- (3, 0);
\draw [thick, -] (3, 0) -- (3.5, 0);\node [right] at (3.5, 0) {Contrastive learning: AVE-Net~\cite{arandjelovic2018objects}, ADC-SSL~\cite{sheng2021cross}...};
\draw [thick, -] (0, -1) -- (0.5, -1);\node [right] at (0.5, -1) {\textbf{Visual Speech Generation~(Section~\ref{sec VSG})}};
\draw [thick, -] (1, -1.5) -- (1, -6);
\draw [thick, -] (1, -2) -- (1.5, -2);\node [right] at (1.5, -2) {Two-stage Framework (Section~\ref{subsec ts vsg})};
\draw [thick, -] (2, -2.5) -- (2,-5);
\draw [thick, -] (2, -3) -- (2.5, -3);\node [right] at (2.5, -3) {Landmark based: ObamaNet~\cite{kumar2017obamanet},ATVG~\cite{chen2019hierarchical}...};
\draw [thick, -] (2, -4) -- (2.5, -4);\node [right] at (2.5, -4) {Coefficient based: AAM~\cite{taylor2017deep}, Blendshape~\cite{thies2020neural}, 3DMM~\cite{zhang2021flow}};
\draw [thick, -] (2, -5) -- (2.5, -5);\node [right] at (2.5, -5) {Vertex based: VOCA~\cite{cudeiro2019capture}, LipsyNc3D~\cite{lahiri2021lipsync3d}...};

\draw [thick, -] (1, -6) -- (1.5, -6);\node [right] at (1.5, -6) {One-stage Framework (Section~\ref{subsec os vsg})};
\draw [thick, -] (2, -6.5) -- (2,-8);
\draw [thick, -] (2, -7) -- (2.5, -7);\node [right] at (2.5, -7) {GAN based: DAVS~\cite{zhou2019talking}, Wav2Lip~\cite{prajwal2020lip}, S2TF~\cite{sun2021speech2talking}...};
\draw [thick, -] (2, -8) -- (2.5, -8);\node [right] at (2.5, -8) {Others: AD-NeRF~\cite{guo2021ad}, DCKs~\cite{ye2022audio}...};
\end{tikzpicture}}
\caption{A taxonomy of representative visual speech recognition and generation methods.}
\label{fig:taxonomy}
\end{figure}
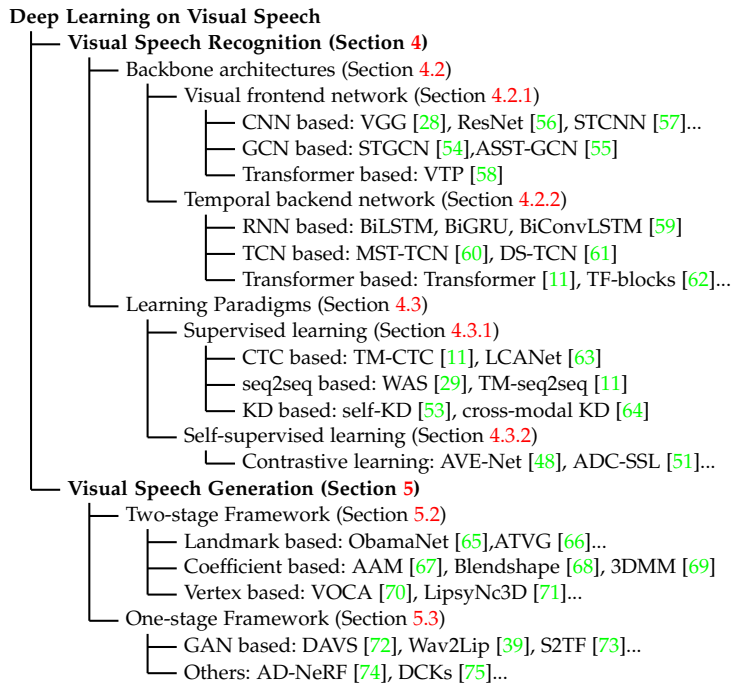

The remainder of this paper is structured as follows. The problem definitions and main challenges of VSA are summarized in Section~\ref{sec Background}.
In Section~\ref{sec data & Eva}, we review the audio-visual datasets and evaluation metrics and compare dataset attributes from multiple perspectives. Section~\ref{sec VSR} illustrates the general framework and representative methods for VSR. Section~\ref{sec VSG} provides a comprehensive survey of existing methods for VSG. A taxonomy of VSR and VSG methods is illustrated in Fig.~\ref{fig:taxonomy}. In Section~\ref{sec conclusion}, we conclude the paper and discuss the possible promising future research directions.

\section{Background}
\label{sec Background}

\begin{figure*}[t]
\centering
\includegraphics[width=0.8\linewidth]{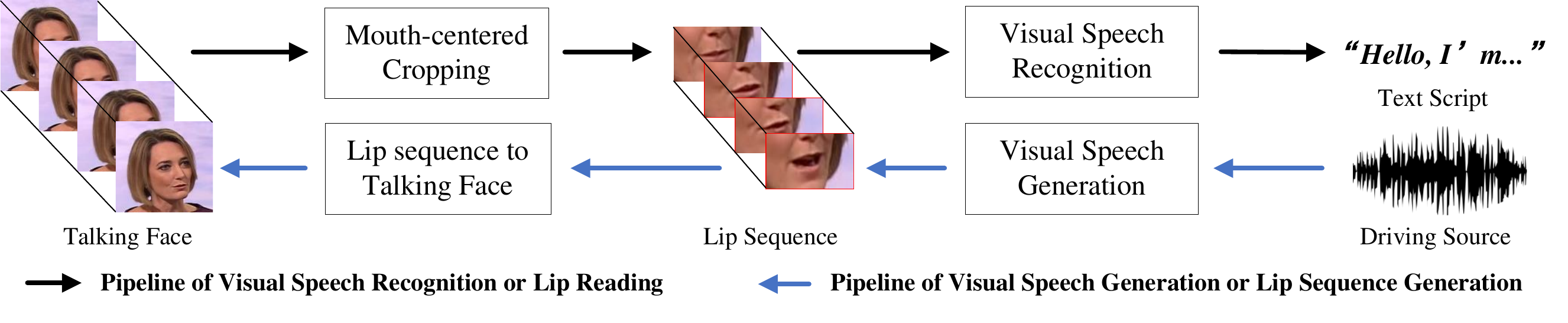} 
\caption{The two formal-dual fundamental problems of visual speech analysis. Top part: Visual speech recognition or lip reading; Bottom part: Visual speech generation or lip sequence generation.}
\label{img tasks}
\end{figure*}

\subsection{The Problems}


Visual speech analysis can be divided into two fundamental problems: recognition and generation. As shown in Fig.~\ref{img tasks}, the two problems are formal-dual and have a reverse pipeline. 

Visual speech recognition (VSR), also known as automatic lip reading, involves designing algorithms to infer the text content according to the speaker's mouth movements. Given a talking face video, a VSR system first crops the video and gets the mouth-centered cropped video. Then, it decodes the cropped video into a specific type of text (words, phrases, or sentences). According to recognition targets, VSR mainly includes two types: word-level and sentence-level. The word-level VSR aims to classify the input video into one of a set of predefined word categories, while the sentence-level VSR tries to predict consecutive sentences from the input video. 

For a better understanding of VSR, we would like to point out the relationship between similar problems: audio speech recognition (ASR), VSR and audio-visual speech recognition (AVSR). All of them share a similar processing pipeline and recognition target, and the main difference lies in the input data type (i.e., audio, lip video, or both). Although audio speech recognition has made significant progress in recent years, efficient utilization of lip video is of great importance in most scenarios, especially when audio signals are corrupted or unavailable~\cite{makino2019recurrent,serdyuk2021audio}. Therefore, the research on VSR is of great significance and applicable.

More specifically, VSR mainly consists of two sub-problems: visual speech representation learning and recognition. The extraction of discriminative visual speech features plays a relatively more important role since even the best recognizer will fail to achieve good results on poor visual speech features.

As a dual task of VSR, the goal of visual speech generation (VSG) is to synthesize a photo-realistic, high-quality talking video that corresponds to the driving source (\eg, a piece of reference audio or text) and the target identity. Specifically, a VSG system first extracts speech representations from the driving source and then fuses the learned speech representations with the target identity to output continual talking frames. 
From the perspective of the learning target, the goal of VSG is more subjective and diverse than that of VSR, making VSG a more challenging problem than VSR.

\begin{figure*}[t]
	\centering
	\includegraphics[width=0.999\textwidth]{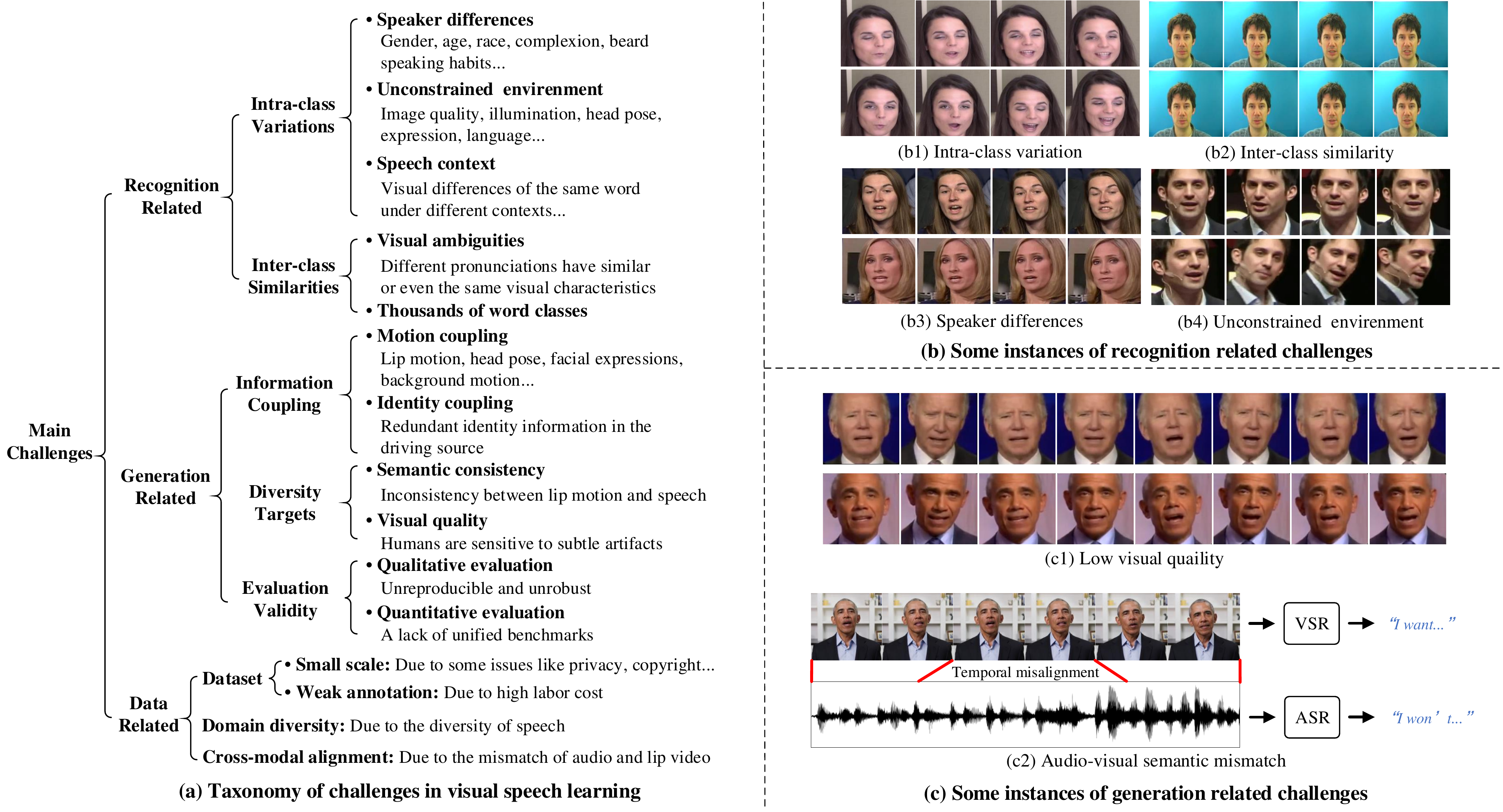}
	\caption{Main Challenges of visual speech analysis. (a) A taxonomy of main challenges. (b) Some practical examples of different challenges. (b1) The upper and lower lines are the respectively different visual dynamics of the word \textit{“wind”} under different contexts; (b2) The upper video instance refers to the word \textit{“place”}, while the lower video refers to the word \textit{"please"}. However, their visual dynamics are very similar; (b3) Two people speak the word \textit{"after"} respectively, with a noticeable difference in their lip motions; (b4) An example of real-time changes in the head pose of a speaker during talking.}
	\label{fig challenges}
\end{figure*}

\subsection{Main Challenges}

\begin{table*}[!t]
\centering
\caption {Statistics of commonly used audio-visual datasets.}
\label{Tab:AV-Datasets}
\begin{threeparttable}
\resizebox{\textwidth}{!}{
\begin{tabular}{|l|c|c|c|c|c|c|c|c|l|c|} 
\hline
Dataset Name & \#Hours    & \#Vocab.                                                      & \#Utter.     & \#Subj. & \begin{tabular}[c]{@{}c@{}}Image size\\~FPS\end{tabular} & Environment    & Data Type                                                                             & Year & \multicolumn{1}{c|}{Highlight}                                                                                                                                                                                                    & Download Link     \\ 
\hline
AVICAR~\cite{lee2004avicar}                                                 & $\sim$33   & \begin{tabular}[c]{@{}c@{}}$26^{\dagger}$\\$13^{\ddagger}$\\1317\end{tabular} & 59k          & 86      & \begin{tabular}[c]{@{}c@{}}720\texttimes 480\\30\end{tabular}       & Car-driving     & \begin{tabular}[c]{@{}c@{}}4-view face- \\ centered videos\end{tabular}                 & 2004 & \begin{tabular}[c]{@{}l@{}}Recorded in a car environment with various noise \\ conditions; Consists of four scripts: isolated digits, \\ isolated letters, phone numbers, and sentences \end{tabular} & \cite{AVICAR}           \\ 
\hline
GRID~\cite{cooke2006audio}                                                & $\sim$28   & 51                                                         & 33k          & 33      & \begin{tabular}[c]{@{}c@{}}720\texttimes 576\\25\end{tabular}       & Lab-controlled & \begin{tabular}[c]{@{}c@{}}3-second face-\\centered videos\end{tabular}               & 2006 & \begin{tabular}[c]{@{}l@{}}Each sentence consists of a six-word~sequence \\of the specific form\end{tabular}                                                                                                                                           & \cite{GRID}             \\ 
\hline
MODALITY~\cite{czyzewski2017audio}                                               & $\sim$31   & 182                                                        & 5880         & 35      & \begin{tabular}[c]{@{}c@{}}1920\texttimes 1080\\100\end{tabular}    & Lab-controlled & \begin{tabular}[c]{@{}c@{}}Stereoscopic \\RGB-D face-\\centered videos\end{tabular} & 2015 & \begin{tabular}[c]{@{}l@{}}Command-like sentences; High resolution with \\ varying noise conditions\end{tabular}                                                                                                                                                 & \cite{MODALITY}         \\ 
\hline
OuluVS2~\cite{anina2015ouluvs2}                                                & $\sim$2    & N/A                                                           & 2120         & 53      & \begin{tabular}[c]{@{}c@{}}1920\texttimes 1080\\30\end{tabular}     & Lab-controlled & \begin{tabular}[c]{@{}c@{}}5-view face- \\ centered videos\end{tabular}                 & 2015 & \begin{tabular}[c]{@{}l@{}}Three types of utterances; High recording quality; \\Five views from 0-90\end{tabular}                                                                                                                                       & \cite{OuluVS2}          \\ 
\hline
IBM AV-ASR~\cite{mroueh2015deep}                                             & $\sim$40   & $\sim$10.4k                                                & N/A          & 262     & \begin{tabular}[c]{@{}c@{}}704\texttimes 480\\30\end{tabular}       & Lab-controlled & \begin{tabular}[c]{@{}c@{}}Face-centered \\videos\end{tabular}                        & 2015 & \begin{tabular}[c]{@{}l@{}}Large vocabulary size; Recorded in clean, studio \\ conditions\end{tabular}                                                                                                                                                  & Unpublic          \\ 
\hline
LRW~\cite{chung2016lip}                                                & $\sim$111  & 500                                                        & $\sim$539K   & 1k+     & \begin{tabular}[c]{@{}c@{}}256\texttimes 256\\25\end{tabular}       & In-the-wild    & \begin{tabular}[c]{@{}c@{}}1.2-second face-\\centered videos\end{tabular}             & 2016 & \begin{tabular}[c]{@{}l@{}}Collected from British television programs; \\Each video corresponds to a word category\end{tabular}                                                                               & \cite{LRW}              \\ 
\hline
LRS2-BBC~\cite{afouras2018deep}                                           & $\sim$225  & $\sim$62.8k                                                & $\sim$144.5k & 1k+     & \begin{tabular}[c]{@{}c@{}}160\texttimes 160\\25\end{tabular}       & In-the-wild    & \begin{tabular}[c]{@{}c@{}}Face-centered \\videos\end{tabular}                        & 2017 & \begin{tabular}[c]{@{}l@{}}Collected from British television; Large-scale; \\Open-world; Sentence-level lip reading\end{tabular}                                                                     & \cite{LRS2}         \\ 
\hline
VoxCeleb1~\cite{nagrani2017VoxCeleb}                                           & $\sim$352  & N/A                                                           & $\sim$153.5k & 1k+     & N/A                                                      & In-the-wild    & \begin{tabular}[c]{@{}c@{}}Public YouTube \\videos\end{tabular}                       & 2017 & \begin{tabular}[c]{@{}l@{}}Large-scale; In-the-wild; Mainly for speaker \\identification and verification\end{tabular}                                                                                                                                             & \cite{VoxCeleb1}        \\ 
\hline
ObamaSet~\cite{suwajanakorn2017synthesizing}                                               & $\sim$14   & N/A                                                           & N/A          & 1       & N/A                                                      & In-the-wild    & \begin{tabular}[c]{@{}c@{}}Public YouTube \\videos\end{tabular}                       & 2017 & \begin{tabular}[c]{@{}l@{}}Focuses on Barack Obama; Collected from Obama's \\weekly presidential addresses; High quality\end{tabular}                                                                           & \cite{ObamaSet}         \\ 
\hline
LRS3-TED~\cite{afouras2018lrs3}                                     & $\sim$475  & $\sim$71.1k                                                & $\sim$151.8k & 5k+     & \begin{tabular}[c]{@{}c@{}}224\texttimes 224\\25\end{tabular}       & In-the-wild    & \begin{tabular}[c]{@{}c@{}}Face-centered \\videos\end{tabular}                        & 2018 & \begin{tabular}[c]{@{}l@{}}Larger-scale; Along with the corresponding \\subtitles and word alignment boundaries\end{tabular}                                                                   & \cite{LRS3}         \\ 
\hline
VoxCeleb2~\cite{chung2018VoxCeleb2}                                              & $\sim$2.4k & N/A                                                           & $\sim$1.1m   & 6k+     & N/A                                                      & In-the-wild    & \begin{tabular}[c]{@{}c@{}}Public YouTube \\videos\end{tabular}                       & 2018 & \begin{tabular}[c]{@{}l@{}}Significantly larger scale; A wide range of different \\ethnicities, accents, professions, and ages\end{tabular}                                                                   & \cite{VoxCeleb2}        \\ 
\hline
LSVSR~\cite{shillingford2018large}                                           & $\sim$3.9k & $\sim$127k                                                 & $\sim$2.9m   & N/A     & N/A                                                      & In-the-wild    & \begin{tabular}[c]{@{}c@{}}Face-centered \\videos\end{tabular}                        & 2018 & \begin{tabular}[c]{@{}l@{}}The largest existing visual speech recognition \\dataset; Extracted from public YouTube videos\end{tabular}                                                                      & Unpublic          \\ 
\hline
LRW-1000~\cite{yang2019lrw}                                & $\sim$57   & 1k                                                            & $\sim$718K   & 2k+     & N/A                                                      & In-the-wild    & \begin{tabular}[c]{@{}c@{}}Mouth-centered \\videos\end{tabular}                       & 2019 & \begin{tabular}[c]{@{}l@{}}The first large-scale Mandarin audio-visual \\speech dataset; collected from TV programs\end{tabular}                                                   & \cite{LRW-1000}         \\ 
\hline
Faceforensics++~\cite{rossler2019faceforensics++}                                        & $\sim$5.7  & N/A                                                           & $\sim$1k     & 1k      & N/A                                                      & In-the-wild    & \begin{tabular}[c]{@{}c@{}}Manipulated \\talking videos\end{tabular}                  & 2019 & Commonly used for facial forgery detection.                                                                                                                                                                   & \cite{Faceforensics++}  \\
\hline
VOCASET~\cite{cudeiro2019capture}                                        & N/A  & N/A                                                           & 255     & 12      & \begin{tabular}[c]{@{}c@{}}5023 vertices\\60\end   {tabular}                                              & Lab-controlled    & \begin{tabular}[c]{@{}c@{}} 3d face \\mesh\end{tabular}                  & 2019 & \begin{tabular}[c]{@{}l@{}}Higher quality 3D scans as well as alignments \\of the entire head\end{tabular}                                                                                                                                    & \cite{VOCASET}  \\
\hline
MEAD~\cite{wang2020mead}                                        & $\sim$39  & N/A                                                           & N/A     & 60      & \begin{tabular}[c]{@{}c@{}}1920\texttimes 1080\\30\end{tabular}                                                      & Lab-controlled    & \begin{tabular}[c]{@{}c@{}}7-view face- \\centered videos\end{tabular}                  & 2020 & Multi-view Emotional Audio-visual Dataset                                                                                                                                                                  & \cite{MEAD}  \\
\hline
HDTF~\cite{zhang2021flow}                                        & $\sim$15.8  & N/A                                                           & 10k+     & 300+      & N/A                                                   & In-the-wild    & \begin{tabular}[c]{@{}c@{}}Face-centered \\ videos\end{tabular}                  & 2021 & \begin{tabular}[c]{@{}l@{}}Higher video resolution than previous in-the-wild \\ datasets\end{tabular}                                                                                                                                                               & \cite{HDTF}  \\
\hline
MuAViC~\cite{anwar2023muavic}                                        & $\sim$1.2k  & N/A                                                           & N/A     & 8k+      & \begin{tabular}[c]{@{}c@{}}224\texttimes 224\\25\end{tabular}                                                   & In-the-wild    & \begin{tabular}[c]{@{}c@{}}Face-centered \\ videos\end{tabular}                  & 2023 & \begin{tabular}[c]{@{}l@{}}The largest open benchmark for multilingual audio-visual  \\ speech recognition\end{tabular}                                                                                                                                                               & \cite{MuAViC}  \\
\hline
\end{tabular}}
\begin{tablenotes}
\item[1] $\dagger$: Alphabets, $\ddagger$: Digits. 
\end{tablenotes}
\end{threeparttable}
\end{table*}

Despite several years of development, most VSA methods have not been capable of meeting real-world requirements due to various challenges. 
As illustrated in Fig.~\ref{fig challenges}(a), to systematically present the challenges in VSA, we classify the main difficulties as recognition-related and generation-related, and discuss the challenges of audio-visual datasets. 

\subsubsection{Recognition-related Challenges}
\label{subsubsec RRC}
From the perspective of representation learning, the ideal of visual speech recognition is to extract speech-related features with strong distinctiveness and robustness. However, both of the two goals suffer from severe practical challenges. Recognition-related challenges mainly stem from (1) the vast range of intra-class variations and (2) the inter-class similarities. 

Intra-class variations can be organized into two types: visual speech intrinsic factors and other recognition-irrelevant factors. In terms of visual speech intrinsic factors, as shown in Fig.~\ref{fig challenges}(b1), a word can produce quite different visual dynamics due to different contexts, continuous reading, speech emotion changing, speaking rate, \etc. Meanwhile, many types of speech-irrelevant interference information dramatically impact visual speech recognition, such as speaker difference, head pose movement, facial expression, imaging condition, and so on. As illustrated in Fig.~\ref{fig challenges}(b3), although the two speakers say the same word ``\textit{after}'', their lip motions look different. 
Nowadays, visual speech recognition in-the-wild has attracted more and more attention, many audio-visual datasets were collected from various realistic scenes (TV News, public speech video...). Fig.~\ref{fig challenges}(b4) shows an example video from the LRS3 dataset, a speaker is talking with dramatic head pose changes. It is hard to eliminate the interference of irrelevant motion information under an unconstrained environment. In addition to head pose changes, illumination, noise corruption, low resolution \etc also bring great difficulties to visual speech recognition.

Besides intra-class variations, visual speech recognition also suffers from inter-class similarities. The phoneme is the most commonly used recognizable unit in a language that distinguishes words from one another. Similarly, the viseme is the smallest recognizable unit of visual speech. There are about three times as many phonemes as visemes in English. Therefore, several phonemes map onto a few visemes. Some phonemes, such as [\textit{p}] and [\textit{b}] [\textit{k}] and [\textit{g}], [\textit{t}] and [\textit{d}], \etc~have almost the same visual characteristics, so they are almost indistinguishable without considering the context in the visual domain. We define this phenomenon as visual ambiguity, the leading cause of inter-class similarities. Fig.~\ref{fig challenges}(b2) demonstrates an instance of word-level visual ambiguity. The other challenge of inter-class similarities stems from thousands of word classes. Subtle differences (\eg~various forms of words) in different word classes make the problem more difficult.

\subsubsection{Generation-Related Challenges}
\label{Generation-Related Challenges}
Different from visual speech recognition, visual speech generation requires not only speech-related information but also identity-related information. As shown in Fig.~\ref{fig challenges}(a), generation-related challenges mainly come from (1) information coupling, (2) diversity targets, and (3) evaluation validity.

A talking face video contains many types of coupled information, such as various motion-related information and identity-related information. For motion coupling, motions that occur on a talking face video can be categorized into two types: intrinsic motions (head pose, facial expression, lip motion,~\etc.) and extrinsic motions (camera motion, background motion,~\etc.). All of these various motions are highly coupled. The motion coupling challenge stems not only from disentangling lip motion from all of these speech-irrelevant motions but also from integrating the synthesized lip sequence into a given identity image. 
The other coupling issue of visual speech generation is identity coupling. As illustrated in Fig.~\ref{fig challenges}(c1), people may feel eerie and uncomfortable while observing these images due to the subtle change of the generated face. This phenomenon, also known as the “uncanny valley effect”, occurs when people observe a synthetic face that's almost human-like but not quite perfect. 
Generally, the driving source contains rich information about the source identity. Therefore, the critical challenge is how to remove the identity information from the driving source to avoid corruption in the process of the target identity synthesis. Besides, most existing methods are only adaptive to specific target identities~\etc. So, the lack of identity generalization is also an important challenge.

Semantic consistency and visual quality are the core desired properties of an excellent VSG method. Semantic consistency represents that the synthesized lip sequence should be synchronized and speech-consistent with the driving source. As shown in Fig.~\ref{fig challenges}(c2), semantic consistency mainly involves two demands: temporal alignment and speech matching. However, the synchronous speech mapping between the driving source and the generated talking video is difficult to realize due to intrinsic differences in temporal resolution and speech characteristics of different data modalities. 
As for visual quality, There are two difficulties: (1) The lack of explicit training objectives since the fidelity and visual quality of the generated lip motion sequences are difficult to define quantitatively. (2) Because of the uncanny valley effect, integrating the generated lip sequence into the whole face with the guidance of people-oriented perceptual is quite challenging.

Besides the aforementioned difficulties, efficiently evaluating visual speech generation methods is another challenge. Existing evaluations, including qualitative and quantitative metrics, have many limitations. For example, qualitative metrics like user study are unreproducible and unstable. As for quantitative metrics, although there are a dozen metrics, some of them are inappropriate and even mutually contradictory.

\subsubsection{Data-Related Challenges}
In addition to the above problem-oriented challenges, audio-visual data-related issues also significantly impact the progress of VSA. Since most of the current deep learning methods are data-driven, the importance of datasets is self-evident. However, existing audio-visual datasets suffer from small scale and weak annotations due to privacy protection and high labor costs.  

The third data-related challenge is domain diversity. Unlike text, visual speech is a biological signal. Its diversity comes from multiple aspects, including language, speaker age, gender, accent, and so on. However, current audio-visual datasets only satisfy partial diversity, which will cause biased training of deep learning models. A potential research direction is to realize cross-modal self-supervised visual speech learning~\cite{sheng2021cross,chung2020perfect,chung2020seeing} based on unlabeled free-access audio-visual data. Despite this, the above issues remain to be resolved.

Another data-related challenge is audio-visual cross-modal alignment. Many deep learning driven technologies~\cite{chung2016out,chung2020perfect, prajwal2020lip, kadandale2022vocalist} have been introduced to address the alignment of A-V signals in a self-supervised way, \ie, automatically creating positive and negative A-V sync/out-of-sync pairs. They usually extracted A-V features from each modality and then measured the similarity/distance between the two features using a sliding window to infer the offset of alignment. However, fine-grained A-V alignment still needs to be further explored due to the temporal-dense nature of speech signals.

\begin{figure*}[t]
	\centering
	\includegraphics[width=0.9\textwidth]{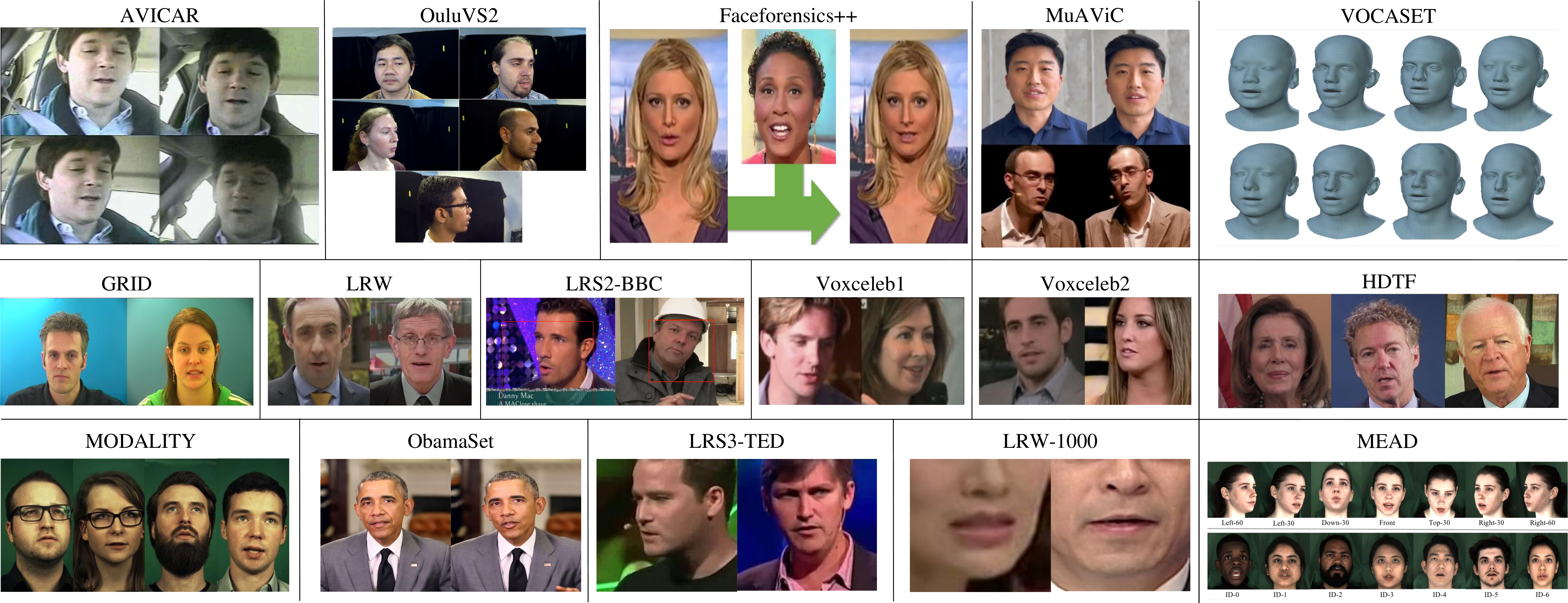}
	\caption{Some example images from AVICAR, OuluVS2, Faceforensics++, GRID, LRW, LRS2-BBC, VoxCeleb1, VoxCeleb2, MODALITY, ObamaSet, LRS3-TED, LRW-1000, VOCASET, HDTF, MEAD, MuAViC. See Table.~\ref{Tab:AV-Datasets} for a summary of these datasets.}
	\label{fig samples}
\end{figure*}

\section{Datasets and Evaluation Metrics}
\label{sec data & Eva}
Datasets have played an important role throughout the history of visual speech research, especially in the big-data era. First, benchmark datasets serve as a common platform for measuring and comparing the performances of competing VSA algorithms. Second, as a typical data-driven learning strategy, deep learning technologies have made significant progress in many audio-visual learning tasks. It is worth noting that the large amounts of annotated data play a crucial role in their success. Third, datasets also further push the field towards increasingly complicated and challenging problems. Therefore, in this section, we first review the existing commonly used datasets for VSA with motivations, statistics, highlights, and download links, then introduce evaluation metrics of different tasks, and finally discuss the future trends in audio-visual datasets.

\subsection{Datasets}
There are dozens of commonly used audio-visual datasets built for VSA. The statistics, highlights and download links are summarized in Table~\ref{Tab:AV-Datasets}, and some selected
sample images are shown in Fig.~\ref{fig samples}. We divide these datasets into two types: controlled and uncontrolled environments. We introduce them briefly in the following.

\subsubsection{Datasets under controlled environments}
As we can see from Table~\ref{Tab:AV-Datasets}, before 2015, visual speech research mainly focused on controlled environments. Controllable factors include recording conditions, equipment, data types, scripts,~\etc. These datasets provide an excellent foundation for visual speech research. 

\textbf{AVICAR}~\cite{lee2004avicar} is the most representative public audio-visual dataset recorded in a car-driving environment. As mentioned above, visual speech can contribute to audio-based speech recognition, especially in noisy environments. Motivated by this, AVICAR is collected for modeling bimodal speech in a driving car, as car driving is a typical acoustic noisy environment.

\textbf{GRID}~\cite{cooke2006audio}, consisting of high-quality audio and video recordings of 1,000 syntactically identical phrases spoken by 34 talkers, is built for comprehensive audio-visual perceptual analysis and microscopic modeling. Besides speech recognition, it can also support audio-visual speech separation tasks.


\textbf{OuluVS2}~\cite{anina2015ouluvs2} is a multi-view audio-visual dataset built for non-rigid mouth motion analysis. It includes 53 speakers uttering three types of utterances. Moreover, it is recorded from five different views spanned between the frontal and profile views. 

\textbf{IBM AV-ASR}~\cite{mroueh2015deep} is a large corpus containing 40 hours of audio-visual recordings from 262 speakers in clean, studio conditions. Compared to previous datasets under controlled environments, it has significant advantages in vocabulary and speaker number. However, this dataset is not publicly available.

\textbf{VOCASET}~\cite{cudeiro2019capture} is a 4D face dataset with about 29 minutes of 4D face scans with synchronized audio from 12 speakers (6 females and 6 males), and the 4D face scans are recorded at 60fps. As a representative high-quality 4D face audio-visual dataset, VOCASET greatly promoted the research on 3D VSG.

\textbf{MEAD}~\cite{wang2020mead}, namely Multi-view Emotional Audio-visual Dataset, is a large-scale, high-quality emotional audio-visual dataset. Unlike previous datasets, it focuses on natural emotional talking face generation and takes multiple emotion states (eight different emotions at three intensity levels) into consideration. 

\subsubsection{Datasets under uncontrolled environments}

Recently, researchers have gradually shifted their focus to in-the-wild visual speech learning. As a result, many large-scale in-the-wild audio-visual datasets are constructed to promote the research. 

\textbf{LRW}~\cite{chung2016lip} is a word-level audio-visual dataset constructed by a multi-stage data automatic collection pipeline. It revolutionarily enlarged the dataset scale and speaker number based on the rich data volume of BBC television programs. It contains over 1,000k word instances spoken by over a thousand people. The main objective of LRW is to test speaker-independent word-level lip reading methods. 

\textbf{LRS2-BBC}~\cite{afouras2018deep} is a sentence-level audio-visual dataset with a similar data collection pipeline and data source as that LRW dataset. It is built for sentence-level lip reading, a more challenging VSR problem than word-level lip reading. All videos in LRS2-BBC are collected from the BBC program, and it contains over 144.5k utterances with a vocabulary size of about 62.8k.

\textbf{VoxCeleb1}~\cite{nagrani2017VoxCeleb} is a large-scale text-independent audio-visual dataset collected from open-source YouTube media. It contains over 100k utterances from 1,251 celebrities. Although it is mainly built for speaker identification, it also can be used for VSG. Its extended version, VoxCeleb2~\cite{chung2018VoxCeleb2}, is currently the largest public available audio-visual dataset.

\textbf{ObamaSet}~\cite{suwajanakorn2017synthesizing} is a specific audio-visual dataset focused on the visual speech analysis of former US President Barack Obama. All video samples are collected from his weekly address footage. Unlike previous datasets, it focuses on Barack Obama only and does not provide any human annotations. 

\textbf{LRS3-TED}~\cite{afouras2018lrs3} is a large-scale sentence-level audio-visual dataset. Compared to LRS2-TED, it has a larger scale in terms of duration, vocabulary, and number of speakers. It consists of talking face videos from over 400 hours of TED and TEDx videos, the corresponding subtitles, and word alignment boundaries. Besides, it is the largest among existing public available annotated English audio-visual datasets.




\textbf{Faceforensics++}~\cite{rossler2019faceforensics++} is an automated benchmark for facial manipulation detection. Different from existing audio-visual datasets, all videos have been manipulated based on DeepFakes~\cite{korshunov2018deepfakes}, Face2Face~\cite{thies2016face2face}, FaceSwap~\cite{FaceSwap}, NeuralTextures~\cite{thies2019deferred} as main methods for facial manipulations. It is commonly used to test forgery video detection methods.

\textbf{HDTF}~\cite{zhang2021flow} is a large-scale in-the-wild audio-visual dataset built for talking face generation. It consists of about 362 different high-resolution videos collected online. Due to the high quality of origin videos, the cropped face-centered videos also have higher visual quality than that of previous datasets like LRW and LRS2-BBC.

\textbf{MuAViC}~\cite{anwar2023muavic} is a multilingual audio-visual corpus for robust speech recognition and robust speech-to-text translation. It consists of 1200 hours of audio-visual speech in 9 languages. The MuAViC dataset is the ﬁrst open benchmark for audio-visual speech-to-text translation and the largest open benchmark for multilingual audio-visual speech recognition.

In addition to the datasets introduced above, there are several audio-visual datasets recorded in different languages. For example, the Chinese Mandarin audio-visual dataset LRW-1000~\cite{yang2019lrw}, Spanish language dataset VLRF~\cite{fernandez2017towards}, Russian language dataset HAVRUS~\cite{verkhodanova2016havrus}~\etc. also promoted the research of VSA on various languages.

Considering that datasets play a crucial role in VSA, we would like to give a summary and discussion of datasets to help readers know the development of VSA.
Compared with early audio-visual datasets, recent ones have improved the number of subjects, dataset scale, recording conditions and script diversity, data quality, \etc. Due to the privacy protection laws (\eg, General Data Protection Regulation (GDPR) in Europe Union), some of the existing large-scale datasets~\cite{mroueh2015deep,shillingford2018large} are not public available. An intuitive solution is to automatically collect available data from online media (\eg, YouTube, BBC, or other online television programs). However, existing audio-visual data auto-collection algorithms may cause a large amount of low-quality data. Therefore, an optimized auto-collection algorithm is crucial for VSA datasets in the future.

\begin{figure*}[t]
	\centering
	\includegraphics[width=0.8\textwidth]{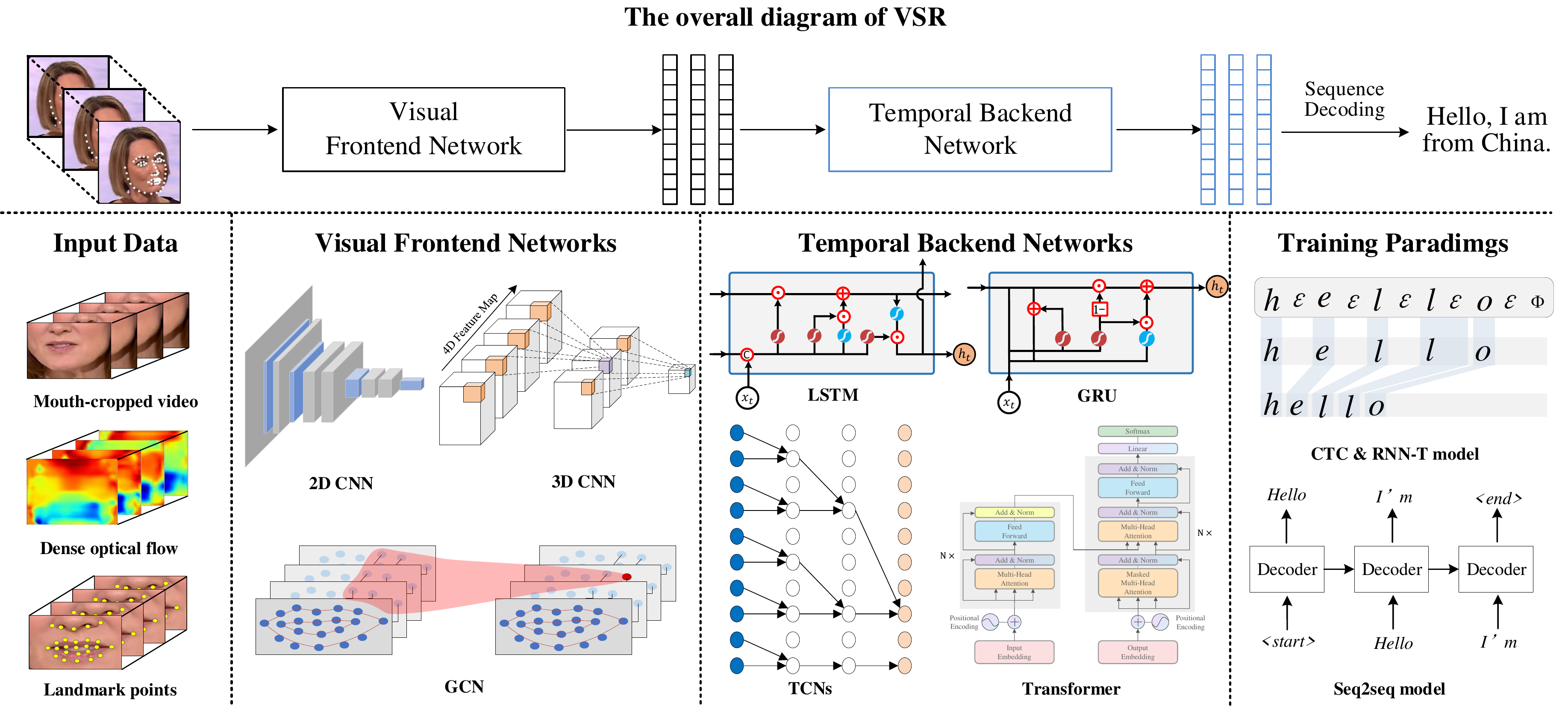}
	\caption{The overall diagram of VSR and various visual frontend networks, temporal backbone networks, and training paradigms. }
	\label{fig vsr framework}
\end{figure*}

\subsection{Evaluation Metrics}
\subsubsection{Evaluation Metrics on VSR}
The word-level VSR task is essentially a multi-class classification problem. Therefore, classification accuracy is the most common evaluation metric for classification models because of its simplicity and efficiency. Besides, $Top-k$ accuracy, namely the standard accuracy of the actual class being equal to any of the $k$ most probable classes predicted by the classification model, is also widely used in VSR.

As for the sentence-level task, Character Error Rate (CER) and Word Error Rate (WER)~\cite{ristad1998learning}, also known as average character-level and word-level edit distances, are the most commonly used evaluation metrics. CER is defined as ${\rm CER} = (S + D + I)/N$, where $S$, $D$ and $I$ are the numbers of substitutions, deletions, and insertions respectively to get from the reference to the hypothesis, and $N$ is the number of characters in the reference. This metric imposes smaller penalties where the predicted string is similar to the ground truth. For example, if the ground truth is ``\textit{about}'' and the model prediction is ``\textit{above}'', then ${\rm CER}=0.4$.
WER and CER are calculated in the same way. The difference lies in whether the formula is applied to character-level or word-level. 

\subsubsection{Evaluation Metrics on VSG}
Appropriate evaluation for VSG remains an open problem, and many recent works have explored various evaluation metrics on VSG. We categorize those metrics based on three learning targets,~\ie, identity preservation, visual quality, and audio-visual semantic consistency. 

\textbf{Identity Preservation.} One of the most important goals of VSG is to preserve the target identity as much as possible during video generation, as humans are quite sensitive to subtle appearance changes in synthesized videos. Since identity is a semantic concept, direct evaluation is not feasible. To evaluate how well the generated video preserves the target identity, existing works usually use the embedding distance of the generated video frames and the ground truth image to measure the identity-preserving performance. For example, Vougioukas~\etal~\cite{vougioukas2018end} adopted the average content distance (ACD)~\cite{tulyakov2018mocogan} to measure the average Euclidean distance of target image representation, obtained using OpenFace~\cite{amos2016openface}, and the representations of generated frames. Besides, Zakharov~\etal~\cite{zakharov2019few} used the cosine similarity between embedding vectors of the ArcFace network~\cite{deng2019arcface} for measuring identity mismatch.

\textbf{Visual Quality.} To evaluate the quality of the synthesized video frames, reconstruction error measurement (\eg, Mean Squared Error) is a natural evaluation way. However, reconstruction error only focuses on pixel-wise alignments and ignores global visual quality. Therefore, existing works usually adopt the Peak Signal-to-Noise Ratio (PSNR) and Structure Similarity Index Measure (SSIM) to evaluate the global visual quality of generated frames. More recently, Prajwal~\etal~\cite{prajwal2020lip} introduced Fr${\rm \acute{e}}$chet Inception Distance (FID) to measure the distance between synthetic and real data distributions, as FID is more consistent with human perception evaluation. Besides, Cumulative Probability Blur Detection (CPBD)~\cite{narvekar2009no}, a non-reference measure, is also widely used to evaluate the loss of sharpness during video generation.

\textbf{Audio-visual Semantic Consistency.} Semantic consistency of the generated video and the driving source mainly contains audio-visual synchronization and speech consistency. For audio-visual synchronization, Landmark Distance (LMD)~\cite{chen2018lip} computes the Euclidean distance of the lip region landmarks between the synthesized video frames and ground truth frames. The other synchronization evaluation metric is to use a pre-trained audio-to-video synchronization network~\cite{chung2019perfect} to predict the offset of generated frames and the ground truth. For the speech consistency, Chen~\etal~\cite{chen2020comprises} proposed a lip-synchronization evaluation metric,~\ie, Lip-Reading Similarity Distance (LRSD), which measures the Euclidean distance of semantic-level speech embeddings obtained by lip reading networks. For better evaluation of speech consistency, lip reading results (accuracy, CER, or WER) comparisons of the generated frames and ground truth are also used as consistency evaluation metrics.

In addition to the above objective metrics, subjective metrics like user study are also widely used in VSG.

\section{Visual Speech Recognition}
\label{sec VSR}

\subsection{The Overall Framework}
Visual Speech Recognition (VSR), also known as lip reading, aims to decode speech from speakers' mouth movements. An essential preprocessing of VSR is mouth-centered region of interest (ROI) cropping. A talking face video contains a large amount of redundant information (such as pose, illumination, gender, skin color,~\etc.) unrelated to the VSR task. To reduce redundant information, it is necessary to crop mouth-centered videos from the raw input video. However, defining the size of mouth-centered ROI is still an open problem. Koumparoulis~\etal~\cite{koumparoulis2017exploring} proved that the selection of ROI will significantly affect the final recognition performance, but it is still unable to determine the optimal ROI.

As shown in Fig.~\ref{fig vsr framework}, a VSR system usually contains three sub-modules.
The first sub-module is visual feature extraction, which intends to extract compact and effective visual feature vectors from mouth-cropped videos. The second sub-module is temporal context aggregation, aiming to aggregate temporal context information for better text script decoding and recognition. The above two sub-modules are also the cores of deep learning based VSR methods. This paper will summarize and discuss existing deep networks for visual feature extraction and temporal context aggregation in Section.~\ref{subsec Backbone}. The last sub-module is text decoding, \ie, converting the feature representations to text. 

The rest of this section is organized as follows. Section.~\ref{subsec Backbone} presents our taxonomy of deep representation learning networks for VSR. Then, we review and discuss various visual speech representation learning paradigms for VSR (supervised learning and unsupervised learning) in Section.~\ref{learning paradigms}. Section.~\ref{VSR Summary} provides a comprehensive summary for readers to know the progress and limitations of existing VSR methods.


\subsection{Backbone Architectures}
\label{subsec Backbone}
Before the era of deep learning, representation learning for VSR had already been explored for a long time. From the feature engineering perspective, traditional feature extraction methods can be categorized into three types: appearance-based, shape-based, and motion-based~\cite{dupont2000audio}. Although simple and explainable, traditional representation learning methods usually do not work well, especially in uncontrolled environments. 
This paper mainly focuses on summarizing and discussing representation learning methods driven by deep learning technologies. Considering the significant difference between deep representation learning and traditional feature extraction, we introduce a novel taxonomy strategy based on two independent parts: visual frontend network and temporal backend network. 



\subsubsection{Visual frontend network}
\label{subsub vfn}
As shown in Fig.~\ref{fig vsr framework}, there are mainly three types of input data: mouth-centered videos, dense optical flow, and landmark points.  Among them, mouth-centered videos and dense optical flow are regular grid data, so CNNs are the most suitable and commonly used backbone architectures for them. On the other hand, as landmark points are irregular data, some existing works~\cite{zhang2021lip,liu2020lip,sheng2021adaptive} adopted Graph Convolution Networks (GCNs) to extract visual features from landmark points. Next, we review these backbone architectures.

\begin{table*}[h]
\centering
\caption{The pros and cons of various visual frontend network architectures and temporal backend network architectures.}
\resizebox{\textwidth}{!}{
\begin{tabular}{|l|l|l|l|}
\hline
Architectures & Available input            & Pros                                                 & Cons                                              \\ 
\hline
2D CNNs       & mouth video                & high memory/time efficiency                          & poor at capturing temporal correlation     \\
3D CNNs       & mouth video / optical flow & powerful at short-term spatiotemporal modeling      & high memory/time cost                        \\
3D + 2D CNNs  & mouth video / optical flow & high memory/time efficiency; strong discrimination & \multicolumn{1}{l|}{not good at capturing subtle lip dynamics}                             \\
Visual transformers          & mouth video        & high robustness; strong discrimination     & high memory/time cost   \\
GCNs          & lip landmark points        & high computation efficiency; semantic preserving     & low accuracy ; low robustness                     \\                  
\hline
RNNs          & \multicolumn{1}{l|}{visual features}      & relatively good generalization                       & short-term dependency; serial computing           \\
Transformers  & \multicolumn{1}{l|}{visual features}      & long-term dependency; parallel computation           & overfitting on small datasets; hard to converge  \\
TCNs          & \multicolumn{1}{l|}{visual features}      & adaptive to multi-scale patterns; high memory efficiency                     & short-term dependency                             \\
\hline
\end{tabular}}
\label{tab backbone}

\end{table*}

\textbf{CNN-based Architectures.} CNNs have been becoming one of the most common architectures in the field of deep learning. Since AlexNet~\cite{Krizhevsky2012} was proposed in 2012, researchers have invented a variety of deeper, wider, and lighter CNN models~\cite{li2021survey}. Representative CNN architectures, such as VGG~\cite{Simonyan2014VGG}, ResNet~\cite{he2016deep}, MobileNet~\cite{Howard2017MobileNets}, DenseNet~\cite{Huang2016Densely}~\etc, have been widely used in learning visual representation for VSR.


The first end-to-end deep visual representation learning for word-level VSR was proposed by Chung~\etal~\cite{chung2016lip}. Based on the VGG-M backbone network, they compared different image sequence input (Multiple Towers \textit{vs.} Early Fusion) and temporal fusion (2D CNNs \textit{vs.} 3D CNNs) architectures and discussed their pros and cons. The experimental results showed that the 2D CNNs are superior to their 3D counterparts by a large margin. However, the above conclusion was not rigorous enough, as the ablation study is insufficient, and word-level VSR datasets have a very short-term dependency. In 2017, Assael~\etal~\cite{assael2016lipnet} proposed LipNet, the first end-to-end sentence-level VSR model. LipNet extracts visual features using a 3-layer Spatiotemporal Convolution Neural Network (STCNN, also known as 3D CNN). The experimental results confirm the intuition that extracting spatiotemporal features using STCNN is better than aggregating spatial-only features.  Considering 3D CNNs are more capable of capturing the dynamics of the mouth region while 2D CNNs are more efficient in time and memory, Stafylakis~\etal~\cite{stafylakis2017combining} proposed to combine 3D CNNs and 2D CNNs for visual feature extraction. Specifically, the proposed visual backbone network consists of a shadow 3D CNN and 2D ResNet. The 3D CNN has just one layer to aggregate short-term temporal information on lip movements. Due to the considerable performance of the model, plenty of VSR models~\cite{petridis2018end,zhang2019spatio,martinez2020lipreading,ma2021towards,sheng2021cross} adopted it as the backbone network for visual features extraction. Recently, Feng~\etal~\cite{feng2020learn} improved this architecture by integrating the Squeeze-and-Extract~\cite{hu2018squeeze} module. Besides VGG and ResNet, researchers have also adopted other representative 2D CNN architectures, including DenseNet~\cite{wang2019multi}, ShuffleNet~\cite{ma2021towards}, MobileNet~\cite{koumparoulis2019mobilipnet},~\etc.

\textbf{GCN-based Architectures.} Considering CNNs are not suitable for irregular grid data, researchers proposed to utilize Graph Convolution Networks (GCNs) to extract visual features from the facial landmark points. Liu~\etal~\cite{liu2020lip} proposed the first end-to-end GCN model (ST-GCN) that extracts shape-based visual features by learning the lip landmark points and their relationships. They first proposed lip graph connection relations and defined the graph adjacency matrices based on the manifold distance of nodes. Then, they combined the image and shape features to extract more discriminative visual features. However, the lip graph connection relations do not naturally exist, and the intuition-guided predefined lip graph restricts the representation ability of shape-based features. Motivated by this, Sheng~\etal~\cite{sheng2021adaptive} proposed an Adaptive Semantic-Spatial-Temporal Graph Convolution Network (ASST-GCN). Unlike~\cite{liu2020lip}, the ASST-GCN parameterizes graph connections and automatically learns adaptive graph connections. Besides, they introduced two graph structures,~\ie, semantic graph and spatial-temporal graph, making graph parameters can be adaptively learned with other parameters in the network training. Existing works show that image-based features are more discriminative than landmark-based features. Sheng~\etal~\cite{sheng2021adaptive} concluded the reason for this. The accuracy of landmark point detection significantly influences its feature discrimination. However, facial landmark detection is challenging, especially in uncontrolled environments.
Since the complementarity between image and landmark features, the combination of CNNs and GCNs is often widely adopted~\cite{zhang2021lip,liu2020lip,sheng2021adaptive}. 

\textbf{Visual Transformer-based Architectures.} Inspired by the significant success of transformer architectures in the field of NLP, researchers have recently applied transformers to computer vision (CV) tasks~\cite{han2022survey}. Recently, transformers have been showing they are potential alternatives to CNNs. Prajwal~\etal~\cite{prajwal2022sub} designed an end-to-end visual transformer-based pooling mechanism that learns to track and aggregate the lip movement representations. The proposed visual backbone network can reduce the need for complicated preprocessing, improving the robustness of visual representation. The ablation study clearly shows that the visual transformer-based pooling mechanism significantly boosts the performance of VSR.

Based on the above backbone architectures, some works further improved visual representation by utilizing two-stream networks. For example, Weng~\etal~\cite{weng2019learning} successfully migrated the two-stream (the raw grayscale video stream and the dense optical flow stream) I3D model to VSR and achieved comparable performance on word-level VSR. However, dense optical flow and 3D convolution calculation is very time-consuming, resulting in low feature extraction efficiency. Wang~\etal~\cite{wang2019multi} utilized 2D CNNs and 3D CNNs to extract both frame-wise spatial features and short-term spatiotemporal features, and then fused the features with an adaptive mask to obtain strong, multi-grained visual features. 

\subsubsection{Temporal backend network}
\label{subsub tbn}
The temporal backend network built upon visual features aims to further aggregate context information. In traditional VSR, classical statistical models (\eg, Hidden Markov Model, HMM) are commonly used for temporal information aggregation. 

\textbf{RNN-based Architectures.} In the field of deep learning, Recurrent Neural Networks (RNNs) are representative network structures used to learn sequence data. The typical RNN structures (\eg, LSTM and GRU) are shown in Fig.~\ref{fig vsr framework}, and their basic structures are similar to that of HMM, in which the dependencies between the observed state sequences are described by the transformation of the hidden state sequence. Compared to HMM, RNNs have a more powerful representation ability due to the nonlinear transformation during hidden state transitions. Bidirectional RNNs (BiRNNs) are variations of basic RNNs, which attempt to aggregate context information from previous timesteps as well as future timesteps. Many works~\cite{assael2016lipnet,chung2017lip,stafylakis2017combining,weng2019learning,feng2020learn,sheng2021cross} have adopted RNN-based (BiLSTM or BiGRU) network architectures as the temporal backend network in VSR. Beyond the above fundamental RNN structures, various modifications~\cite{wang2019multi,chen2020lipreading} have been made to improve feature learning for VSR. For example, Wang~\etal~\cite{wang2019multi} utilized BiConvLSTM~\cite{shi2015convolutional} as temporal backend network. ConvLSTM is a convolutional counterpart of conventional fully connected LSTM, which models temporal dependency while preserving spatial information. Wang~\etal integrated the attention mechanism into the model to further improve the BiConvLSTM architecture. 

\textbf{Transformer-based Architectures.} Compared to RNN-based architectures, Transformers~\cite{vaswani2017attention} have significant advantages in long-term dependency and parallel computation. However, transformers usually suffer from some drawbacks. First, transformers are more prone to overfitting than RNNs and TCNs in small-scale datasets. Second, transformers are limited in some specific tasks (\eg, word-level VSR tasks) with short-term context. Therefore, transformers are more suitable for sentence-level VSR tasks than word-level VSR tasks. ~\cite{afouras2018deep} is the first work introducing transformers to VSR. Based on the basic transformer architecture, the authors proposed two types of backend models: TM-seq2seq and TM-CTC. The difference between the two models lies in the training target. The experiments clearly showed that the transformer-based backend network performs much better than the RNN-based backend network in the sentence-level VSR task. Since the basic transformer pays no extra attention to short-term dependency, Zhang~\etal~\cite{zhang2019spatio} proposed multiple Temporal Focal blocks (TF-blocks), helping features to look around their neighbors and capture more short-term temporal dependencies. The results demonstrated that the short-term dependency is as crucial as the long-term dependency in sentence-level VSR.

\textbf{TCN-based Architectures.} In the context of deep sequence models, RNNs and Transformers have a high demand for memory and computation ability. Temporal Convolutional Networks (TCNs) are another type of deep sequence model, and various improvements have been applied to the basic TCN to make them more appropriate for VSR. For example, Afouras~\etal~\cite{afouras2018deep_a} used depth-wise separable convolution (DS-TCN) for sentence-level VSR. However, the performance of DS-TCN does not work as well as transformers, as TCN-based models have a poor ability to capture long-term dependency. To enable temporal backend network capturing multi-scale temporal patterns, Martinez~\cite{martinez2020lipreading} proposed to utilize a multi-scale TCN (MS-TCN) structure, which achieved SOTA results (87.9\% Acc) on the word-level LRW dataset.

Table.~\ref{tab backbone} summarizes the general pros and cons of various visual frontend networks and temporal backend networks, and the available inputs of the corresponding visual frontend network. As we know, most of the existing VSR models are derived from general backbone models used in other fields (\eg, action recognition~\cite{carreira2017quo,yan2018spatial}, audio speech recognition~\cite{shi2015convolutional},~\etc.), and few are designed explicitly for VSR. Therefore, more attention should be paid to the particular structure adaptive to the properties of VSR in the future.


\subsection{Learning Paradigms}
\label{learning paradigms}

\subsubsection{Supervised learning for VSR}
\label{sl for vsr}
There are two mainstream VSR tasks: word-level and sentence-level. With a limited number of word categories, the former is to recognize isolated words from the input videos (\ie, talking face video classification), usually trained with multi-classification cross-entropy loss. The latter is to make unconstrained sentence-level sequence prediction. However, due to the unconstrained word categories and video frame length, it is much more complicated than the word-level VSR task.

Supervised learning of end-to-end sentence-level VSR tasks (sentence prediction) can be divided into two types. Given the input sequence, the first type uses a neural network as an emission model, which outputs the likelihood of each output symbol (\eg, phonemes, characters, words). A popular version of this variant is the Contortionist Temporal Classification (CTC)~\cite{graves2006connectionist}, where the model predicts frame-wise labels and then looks for the optimal alignment between the frame-wise predictions and the output sequence. The main weakness of CTC is that the output labels are not conditioned on each other (it assumes each unit is conditional independent), and hence, a language model is needed as a post-processing step. Another core assumption of CTC is that it assumes a monotonic ordering between input and output sequences, which is suitable for VSR but not for machine translation. To overcome the conditional independent issue of CTC, a commonly used modified version of the CTC model is the RNN-Transducer (RNN-T)~\cite{graves2013speech} in speech recognition. RNN-T combines a CTC-like paradigm with a separate RNN that predicts each output given the previous ones, thereby yielding a jointly trained acoustic and language model. RNN-T solves the problem of insufficient language modeling ability in the CTC model. Despite this, RNN-T suffers from slow convergence speed in training and difficulty in effectively parallel training.

The second type is sequence-to-sequence (seq2seq) models that first read the whole input sequence before predicting the output sentence. 
Chan~\etal~\cite{chan2016listen} proposed an elegant seq2seq method to transcribe audio signals to characters. Seq2seq models decode an output symbol at time $t$ (\eg, phonemes, characters, words) conditioned on previous outputs $1, . . . , t-1$. Thus, unlike CTC-based models, the model implicitly learns a language model over output symbols, and no further processing is required. However, it has been shown~\cite{kannan2018analysis} that it is beneficial to incorporate an external language model in the decoding of seq2seq models as well. Chung~\etal~\cite{chung2017lip} proposed the WAS (Watch, Attend and Spell) model, which is a classical seq2seq VSR model. With the help of the attention mechanism, the WAS model is more capable of capturing long-term dependency.

Based on the transformer backbone architecture, Afouras~\etal~\cite{afouras2018deep} have deeply analyzed the pros and cons of the CTC model and the seq2seq model for VSR. Generally, the seq2seq model performs better than the CTC model in the sentence-level VSR task. However, the seq2seq model needs more training time and inference time. Besides, the CTC model generalizes better and adapts faster as the sequence lengths are increased.

Besides the above label-level supervised learning paradigms, feature-level supervised learning is also widely explored in VSR. Knowledge distillation (KD) technology is the key to feature-level supervised learning. For example, Ma~\etal~\cite{ma2021towards} proposed a multi-stage self-KD training framework for the word-level VSR task. Like label smoothing, KD can provide an extra supervisory signal with inter-class similarity information. Some works~\cite{afouras2020asr,li2019improving,yu2020audio} utilized cross-modal KD to train a robust VSR model by distilling from a well-trained ASR model. With the help of the ASR model, KD technology can significantly speed up the training of the VSR model. Meanwhile, combining CTC loss and KD loss can further improve the performance of VSR. 


\subsubsection{Unsupervised learning for VSR}
\label{ul for vsr}
Unsupervised learning for VSR aims to learn discriminative visual representations without access to manual annotation. Among multiple unsupervised learning frameworks, cross-modal self-supervised learning is dominant in VSR.
Despite the remarkable progress witnessed in the past decade, the successes of supervised deep learning rely heavily on vast manually annotated training data, which has severe limitations in many real-world applications, including the VSR task. Firstly, supervised learning is restricted to relatively narrow domains primarily defined by the labeled training data and thus leads to limited generalization ability. Secondly, a large amount of accurately labeled data, like a large-scale annotated dataset for VSR, is costly to gather. Recently, self-supervised learning has received growing attention due to its high label efficiency and good generalization.

\begin{table*}
\centering
\caption{The comparison of representative VSR methods.}
\label{SOTA comparison}
\resizebox{\linewidth}{!}{%
\begin{tabular}{|c|c|c|c|c|c|c|c|c|c|} 
\hline
\multicolumn{1}{|c|}{\multirow{2}{*}{Task type}} & \multirow{2}{*}{Method}             & \multirow{2}{*}{Frontend network}                               & \multirow{2}{*}{Backend network}                                  & \multicolumn{4}{c|}{Experimental settings}   & \multirow{2}{*}{Performance (Dataset)}     & \multirow{2}{*}{Highlights}    \\ 
\cline{5-8}
\multicolumn{1}{|c|}{}   &    &     &    & Learning paradigm    & Extra datasets   & Extra LM    & output symbol   &  &   \\ 
\hline
\multirow{21}{*}{\rotatebox{90}{Word-level}}    & Chung~\etal~\cite{chung2016lip}    & VGG-M     & /    & /    & /   & /     & word      & \begin{tabular}[c]{@{}c@{}}61.1\% (LRW)\\25.7\%(LRW-1000)\end{tabular}  & \begin{tabular}[c]{@{}c@{}}Discussed various temporal fusion ways for \\ word-level VSR networks\end{tabular}    \\ 
\cline{2-10}  & 
Stafylakis~\etal~\cite{stafylakis2017combining}               & C3D-ResNet34   & BiLSTM     & /    & /     & /                    & word                       & \begin{tabular}[c]{@{}c@{}}83.5\% (LRW)\\38.2\% (LRW-1000)\end{tabular}                         & \begin{tabular}[c]{@{}c@{}}Proposed the most widely used visual \\ frontend network,~\ie, C3D\_ResNet \end{tabular}                   \\ 
\cline{2-10}  & 
Wang~\etal~\cite{wang2019multi}  & \begin{tabular}[c]{@{}c@{}}ResNet34\\3D-DenseNet52\end{tabular} & BiConvLSTM   & /    & /  & /  & word  & \begin{tabular}[c]{@{}c@{}}83.3\% (LRW)\\36.9\% (LRW-1000)\end{tabular}  & \begin{tabular}[c]{@{}c@{}}Introduced the BiConvLSTM architecture \\ for VSR\end{tabular}   \\ 
\cline{2-10}  & 
Liu~\etal~\cite{liu2020lip}  & \begin{tabular}[c]{@{}c@{}}C3D-ResNet34\\ST-GCN\end{tabular}    & BiGRU   & /    & /    & /                    & word    & 84.25\% (LRW)    & \begin{tabular}[c]{@{}c@{}}Firstly utilized the GCN-based network \\ in VSR\end{tabular}          \\ 
\cline{2-10} & 
Martinez~\etal~\cite{martinez2020lipreading}  & C3D-ResNet18   & MS-TCN  & /   & /   & /    & word     & \begin{tabular}[c]{@{}c@{}}85.3\% (LRW)\\41.4\% (LRW-1000)\end{tabular}  & \begin{tabular}[c]{@{}c@{}}Improved performance by multi-scale TCN;\\Adaptive to varying input length\end{tabular}                                             \\ 
\cline{2-10} & 
Sheng~\etal~\cite{sheng2021adaptive}    & \begin{tabular}[c]{@{}c@{}}C3D-ResNet18\\ASST-GCN\end{tabular}  & MSTCN   & /  & /  & /    & word    & 85.7\% (LRW)   & \begin{tabular}[c]{@{}c@{}}Introduced lip semantic encoding;~\\No need for predefined lip graph\end{tabular}  \\ 
\cline{2-10} & 
Ma~\etal~\cite{ma2021towards}   & C3D-ResNet18   & MS-TCN   & \begin{tabular}[c]{@{}c@{}}Multi-stage \\KD\end{tabular}    & /    & /   & word   & \begin{tabular}[c]{@{}c@{}}\textbf{87.7\%} (LRW)\\43.2\% (LRW-1000)\end{tabular}   & \begin{tabular}[c]{@{}c@{}}Improved generalization with the help of KD \\and achieved SOTA results on LRW\end{tabular}   \\ 
\cline{2-10}  & 
Feng~\etal~\cite{feng2020learn}   & SE-C3D-ResNet18   & BiGRU   & /      & /    & /   & word     & \begin{tabular}[c]{@{}c@{}}85.0\% (LRW)\\\textbf{48.0\%} (LRW-1000)\end{tabular}                         & \begin{tabular}[c]{@{}c@{}}Introduced Squeeze-and-Extract module;\\Achieved SOTA results on LRW-1000\end{tabular}   \\ 
\cline{2-10}     & 
Yang~\etal~\cite{yang2022cross}    & C3D-ResNet18  & ResNet18             & \begin{tabular}[c]{@{}c@{}}Cross-modal\\mutual learning\end{tabular} & /   & /   & word    & \begin{tabular}[c]{@{}c@{}}$88.5\%^{\dagger}$ (LRW)\\$\mathbf{50.5\%^{\dagger}}$ (LRW-1000)\end{tabular}                         & \begin{tabular}[c]{@{}c@{}}Proposed a unified framework for audio-\\visual speech recognition and synthesis\end{tabular}   \\ 
\cline{2-10}     & 
Koumparoulis~\etal~\cite{koumparoulis2022accurate}    & EfficientNetV2-L
  & TCN + Transformer   & / & /   & /   & word    & $\mathbf{89.5\%} $(LRW)      & \begin{tabular}[c]{@{}c@{}}Proposed a resource-efﬁcient network for \\visual speech recognition\end{tabular}   \\

\hline
\multirow{23}{*}{\rotatebox{90}{Sentence-level}}              & Assael~\etal~\cite{assael2016lipnet}                 & ST-CNN                                                          & BiGRU                                                             & CTC loss                                                          & /                                                                           & $\surd$                  & character                  & \begin{tabular}[c]{@{}c@{}}1.9\% CER (GRID)\\4.8\% WER (GRID)\end{tabular}                      & \begin{tabular}[c]{@{}c@{}}The first end-to-end~sentence-level VSR model\end{tabular}                                                                         \\ 
\cline{2-10}
                                              & Xu~\etal~\cite{xu2018lcanet}                     & C3D + HighwayNet                                                & BiGRU                                                             & CTC loss                                                          & /                                                                           & /                    & character                  & \begin{tabular}[c]{@{}c@{}}1.3\% CER (GRID)\\2.9\% WER (GRID)\end{tabular}                      & \begin{tabular}[c]{@{}c@{}}Compensated the defect of the CTC approach\end{tabular}                                                                              \\ 
\cline{2-10}
                                              & \multirow{3}{*}{Afouras~\etal~\cite{afouras2018deep}} & \multirow{3}{*}{C3D-ResNet18}                                   & \multirow{3}{*}{Transformer}                                      & CTC loss                                                          & \multirow{4}{*}{\begin{tabular}[c]{@{}c@{}}LRW, MVLRS, \\LRS2\end{tabular}} & \multirow{3}{*}{$\surd$} & \multirow{3}{*}{character} & \begin{tabular}[c]{@{}c@{}}54.7\% CER (LRS2)\\66.3\% WER (LRS3)\end{tabular}                    & \multirow{3}{*}{\begin{tabular}[c]{@{}c@{}}Deeply analyzed the pros and cons of the CTC \\ model and the seq2seq model~\end{tabular}}  \\ 
\cline{5-5}\cline{9-9}
                                              &                                     &                                                                 &                                                                   & seq2seq loss                                                      &                                                                             &                      &                            & \begin{tabular}[c]{@{}c@{}}48.3\% CER (LRS2)\\58.9\% WER (LRS3)\end{tabular}                    &                                                                                                                                                                 \\ 
\cline{2-10}
                                              & Shillingford~\etal~\cite{shillingford2018large}           & ST-CNN                                                          & BiLSTM                                                            & CTC loss                                                          & LSVSR                                                                       & $\surd$                  & phoneme                    & \begin{tabular}[c]{@{}c@{}}28.3\% CER (LSVSR)\\40.9\% WER (LSVSR)\\55.1 WER (LRS3)\end{tabular} & \begin{tabular}[c]{@{}c@{}}Adopted phoneme as the output symbol \\and proposed the largest dataset LSVSR\end{tabular}                             \\ 
\cline{2-10}
                                              & Zhang~\etal~\cite{zhang2019spatio}                   & C3D-ResNet18                                                    & TF-blocks                                                         & seq2seq loss                                                      & LRW, LRS2                                                                   & /                    & character                  & \begin{tabular}[c]{@{}c@{}}1.3\% WER (GRID)\\51.7\% WER (LRS2)\\60.1\% WER (LRS3)\end{tabular}  & \begin{tabular}[c]{@{}c@{}}Integrated causal convolution into transformer \\ for VSR\end{tabular}                                                                        \\ 
\cline{2-10}
                                              & Makino~\etal~\cite{makino2019recurrent}                 & ST-CNN                                                          & RNN-T (BiLSTM)                                                    & seq2seq loss                                                      & YT-31khrs                                                                   & $\surd$                  & character                  & 33.6\% WER (LRS3)                                                                               & \begin{tabular}[c]{@{}c@{}}Proposed an RNN-T based VSR system and \\ collectd a large dataset from  YouTube\end{tabular}                         \\ 
\cline{2-10}
                                              & Ma~\etal~\cite{ma2021end}                      & C3D-ResNet18                                                    & \begin{tabular}[c]{@{}c@{}}Conformer +\\Transformer~\end{tabular} & \begin{tabular}[c]{@{}c@{}}CTC loss + \\seq2seq loss\end{tabular} & \begin{tabular}[c]{@{}c@{}}LRW, LRS2,\\LRS3\end{tabular}                    & $\surd$                  & character                  & \begin{tabular}[c]{@{}c@{}}37.9\% WER (LRS2)\\43.3\% WER (LRS3)\end{tabular}                    & \begin{tabular}[c]{@{}c@{}}Proposed a hybrid CTC/Attention model \\ and achieved SOTA results on LRS3\end{tabular}                           \\ 
\cline{2-10}
                                              & Prajwal~\etal~\cite{prajwal2022sub}                & 3DCNN+VTP                                                       & Transformer                                                       & seq2seq loss                                                      & \begin{tabular}[c]{@{}c@{}}LRS2, LRS3,\\MVLRS, TEDx\end{tabular}            & $\surd$                  & sub-word                  & \begin{tabular}[c]{@{}c@{}}\textbf{22.6\%} WER (LRS2)\\\textbf{30.7\%} WER (LRS3)\end{tabular}                    & \begin{tabular}[c]{@{}c@{}}Introduced sub-word as the output symbol and \\ replaced the 2DCNN with the visual transformer\end{tabular}                \\
\hline
\end{tabular}
}
\begin{tablenotes}
\item[1] $\dagger$: audio data is used in the training. 
\end{tablenotes}
\end{table*}

Recent advances in cross-modal self-supervised learning have shown that the corresponding audio can serve as a supervisory signal to learn effective visual representations for VSR. Audio-visual self-supervised learning aims to extract efficient representations from the co-occurring A-V data pairs without extra annotation. Based on the natural synchronization property of audio and video, existing methods mainly adopt contrastive learning to achieve this goal. Chung~\etal~\cite{chung2016out} are the first to train an A-V synchronization model in an end-to-end manner with margin-based~\cite{hadsell2006dimensionality} pairwise contrastive loss. Besides VSR, they have proved that the trained network work can effectively be finetuned to other tasks like speaker detection. With the same training strategy, Korbar~\etal~\cite{korbar2018cooperative} broadened the scope of the study to encompass arbitrary human activities rather than lip movements. Except for margin-based loss, L1 loss and binary classification loss~\cite{arandjelovic2018objects,owens2018audio,senocak2018learning,ma2021lira} are also widely used for A-V representations learning. Those works have proved the learned A-V representations can be further transferred to more downstream tasks, such as visualizing the locations of sound sources, action recognition, audio-visual source separation, \etc Recently, Chung~\etal~\cite{chung2020perfect} reformulated the contrastive task as a multi-way matching task and demonstrated that using multiple negative samples can improve the performance. Considering existing methods only exploit the natural synchronization of the video and the corresponding audio, Sheng~\etal~\cite{sheng2021cross} proposed a novel self-supervised learning framework called Adversarial Dual-Contrast Self-Supervised Learning (ADC-SSL), to go beyond previous methods by explicitly forcing the visual representations disentangled from speech-unrelated information. To achieve this goal, they combine contrastive learning and adversarial training through three pretext tasks: A-V synchronization, identity discrimination, and modality classification. 

\subsection{Summary and performance comparison}
\label{VSR Summary}
We have witnessed significant progress in various aspects of visual speech recognition. In this subsection, we will compare existing VSR methods on representative datasets and summarize the main issues of VSR. 

\subsubsection{Performance Comparison}
In this section, we compare the existing deep learning-based VSR methods. Due to the large number of methods proposed for VSR, it is not possible to list and compare all of them. Thus, we select representative works and several milestone methods. Table.~\ref{SOTA comparison} summarizes the performance and some experimental settings of some representative VSR methods on large-scale commonly used benchmark datasets, including LRW~\cite{chung2016lip}, LRW-1000~\cite{yang2019lrw}, GRID~\cite{cooke2006audio}, LRS2~\cite{afouras2018deep} and LRS3~\cite{afouras2018lrs3}.

As for the word-level VSR task, various visual frontend networks have been designed to boost the performance, such as VGG-M, C3D-ResNet, ST-GCN, ASST-GCN,\etc. Among them, the C3D-ResNet architecture is the most widely used.~\cite{stafylakis2017combining} provided the baseline (C3D-ResNet34 + BiLSTM, 83.5\%) on the LRW dataset. Many subsequent works inherited this structure and further improved the performance by introducing some tricks, such as label smoothing, weight decay, dropout, Squeeze-and-Extract module, two-stream, multi-stage KD,~\etc. As for temporal backend networks, RNN-based models and TCN-based models have similar performance. Based on C3D-ResNet18 + MSTCN, Ma~\etal~\cite{ma2021towards} improved the SOTA to 87.7\% on LRW.
Recently, more and more works~\cite{chung2016out,arandjelovic2018objects,chung2020perfect, chung2020seeing, sheng2021cross, yang2022cross} tried to improve visual representations by utilizing extra audio information in the training stage rather than the design of network architectures, as audio signals can provide more fine-grained supervision than text annotations. The SOTA results (88.5\% on LRW and 50.5\% on LRW-1000) were realized based on cross-modal audio-visual mutual learning~\cite{yang2022cross}. 

As for the sentence-level VSR task, deep learning-based VSR methods have vastly outperformed human lip-readers~\cite{assael2016lipnet}. As shown in Table.~\ref{SOTA comparison}, deep VSR models almost reach performance saturation (SOTA result: 1.3\% WER) on the simple (constrained recording environment and limited corpus) GRID dataset. Therefore, researchers pay more attention to VSR in unconstrained environments. Motivated by the practical need, we focus more on the large-scale in-the-wild datasets (\eg, LRS2 and LRS3). The fair performance comparison of sentence-level VSR methods is quite hard, as there are too many extra influencing factors. For example, some methods trained the model with extra datasets (Some of them are not public available). Besides, the outputs of the model are generally optimized by extra language models, while language models are trained with existing large-scale text corpus. The introduction of language models can significantly improve performance, and it is not fair to compare these methods optimized by different language models. Therefore, to make it clearer for readers, we list some representative sentence-level VSR models and their experimental settings in Table.~\ref{SOTA comparison}. 

\subsubsection{Main issues and facts}
Over the last decade, deep learning-based VSR techniques have been significantly developed. However, some issues remain to be solved. We conclude them as follows:
\begin{itemize}
\renewcommand{\labelitemi}{$\bullet$}
    \item The cropping preprocessing of raw talking face videos have a significant impact on the recognition results, and how to define the optimal lip ROI for the VSR task is worthy of further exploration.
    \item In practical applications, real-time is another substantial demand for VSR. However, most existing VSR methods only focus on recognition accuracy but ignore real-time. Therefore, the trade-off between accuracy and real-time should be considered in the future.
    \item There is no formal robustness analysis of existing VSR methods. As we have mentioned in Section.~\ref{subsubsec RRC}, VSR faces many challenges, such as speaker differences and unconstrained environments. Existing deep learning-based VSR networks are rarely targeted to solve these problems. Therefore, the robustness analysis of VSR methods needs more attention in the future.
    \item Another serious problem of VSR research is the lack of fair benchmarks for algorithm comparison, especially for the sentence-level VSR task. The performance of VSR is affected by many factors, such as extra language models, multiple training datasets, audio signals, and implementation details. Due to the lack of a unified test platform, a fair comparison of VSR algorithms is not easy to achieve.
\end{itemize}

\section{Visual Speech Generation}
\label{sec VSG}

Visual Speech Generation (VSG), also known as lip sequence generation, aims to synthesize a lip sequence corresponding to the driving source (a clip of audio or a piece of text).

Traditional VSG methods suffer from severe practical challenges~\cite{mattheyses2015audiovisual}, such as complex generation pipelines, constrained applicable environments, over-reliance on fine-grained viseme (phoneme) annotations,~\etc. To realize mapping driving sources to lip dynamics, representative traditional VSG methods mainly adopted cross-modal retrieval approaches~\cite{bregler1997video,garrido2014automatic,garrido2015vdub,thies2016face2face} and HMM-based approaches~\cite{fu2005audio,xie2007realistic}. For example, Thies~\etal~\cite{thies2016face2face} introduced a typical image-based mouth synthesis approach that generates a realistic mouth interior by retrieving and warping best-matching mouth shapes from offline samples. However, retrieval-based methods are static text-phoneme-viseme mappings and do not really consider the contextual information of the speech. Meanwhile, retrieval-based methods are pretty sensitive to head pose changes. HMM-based methods also suffer from some drawbacks, such as the limitation of the prior assumptions (\eg, Gaussian Mixture Model (GMM) and its diagonal covariance). As deep learning technologies have extensively promoted the developments of VSG, we focus on reviewing deep learning based VSG methods in this section.

To make the scope of VSG clear for readers, we first explain the relationship and difference between VSG and another hot topic,~\ie, Talking Face Generation (TFG)~\footnote{Talking Face Generation is also called talking face synthesis, talking head generation, or talking portraits generation. These concepts are interchangeable, and to be consistent, the expression ``Talking Face Generation (TFG)'' is adopted in this paper.}~\cite{jamaludin2019you,zhou2019talking}.


TFG aims to synthesize a realistic, high-quality talking face video corresponding to the driving source and the target identity. According to the modality of driving sources, TFG can be divided into audio-driven, text-driven, and video-driven. Among them, video-driven TFG mainly focuses on video-oriented face-to-face facial expression transferring rather than visual speech generation. Therefore, video-driven TFG methods will not appear in this paper. 

Traditionally, VSG can be viewed as a key sub-component of text-driven (audio-driven) TFG. The other component is video editing, following a specific editing pipeline to output the final synthesized talking face video based on the generated lip sequence. Recently, to reduce manual intervention and simplify the complexity of the overall pipeline, more and more researchers have tried to synthesize full talking faces in an end-to-end manner instead of lip sequence. Consequently, the definition boundary between VSG and text-driven (audio-driven) TFG is getting blurred, which means some text-driven (audio-driven) TFG methods are also in our review scope. Therefore, to give a comprehensive survey on VSG, we also review some TFG methods driven by text and audio, as these works implicitly or explicitly involve VSG modules.

\subsection{The Overall Pipeline}

Given a reference identity (an image or a 3D facial model of the target speaker) and a driving source (a piece of audio or text), the objective of VSG is to generate the final synthesized talking lip (face) videos. Existing VSG approaches have various properties, such as input modalities (text-driven or audio-driven), synthesizing strategies (computer graphics based, image reconstruction based, or hybrid based), speaker generalization (speaker-independent or speaker-dependent), learning paradigms (supervised learning or unsupervised learning), classifying these approaches is not an easy task. 

\begin{figure}[t]
	\centering
	\includegraphics[height=0.81\textheight]{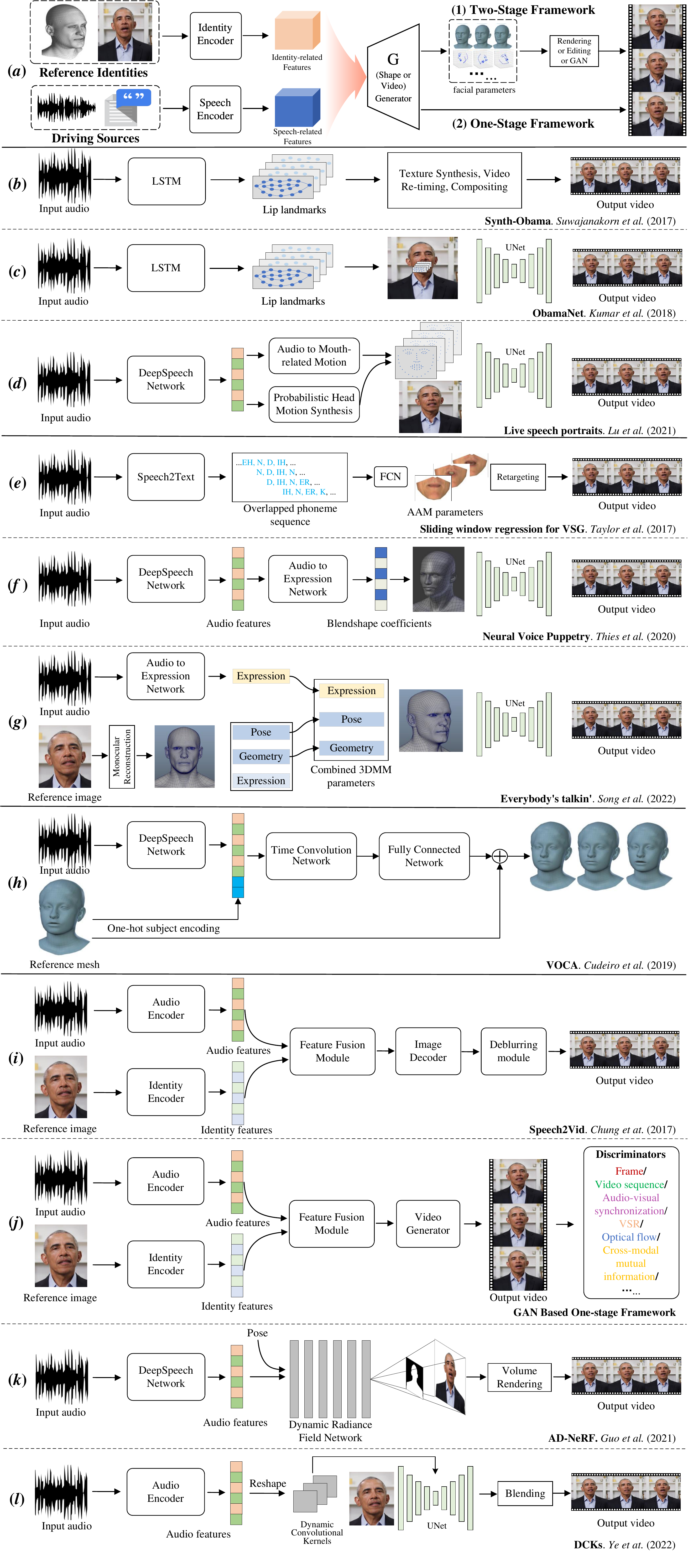}
	\caption{\textit{(a)}: The overall framework of visual speech generation. \textit{(b)-(h)}: Representative Two-stage VSG methods. \textit{(i)-(l)}: Representative One-stage VSG methods.}
	\label{fig deep VSG}
\end{figure}

This section provides a novel taxonomy for VSG methods, as shown in Fig.~\ref{fig deep VSG}(a). In specific, we organize VSG approaches into two frameworks: a) Two-stage frameworks, which include two mapping steps,~\ie, driving source to facial parameters and facial parameters to videos; and b) One-stage (Unified) frameworks, having a single generation process which is intermediate facial parameters free. Next, we review and analyze current two-stage and one-stage VSG methods as well as their advantages and disadvantages in detail in Section.~\ref{subsec ts vsg} and~\ref{subsec os vsg}, respectively. 


\subsection{Two-Stage VSG Framework}
\label{subsec ts vsg}
The two-stage VSG frameworks mainly consist of two steps: a) mapping the driving source to facial parameters using DNNs and b) transforming the learned facial parameters to output videos based on GPU rendering, video editing, or Generative Adversarial Networks (GANs)~\cite{goodfellow2014generative}.
According to the data type of facial parameters, existing two-stage VSG approaches can be divided into Landmarks based, Coefficients based, Vertex based, and others. 

\subsubsection{Landmark based Methods}
Facial landmark points around facial components capture the rigid and non-rigid facial deformations due to head movements and facial expressions~\cite{wu2019facial}. Facial landmark points are widely used in various facial analysis tasks, including VSG. As a pioneering work, Suwajanakorn~\etal~\cite{suwajanakorn2017synthesizing} adopted a simple single-layer LSTM with the time delay mechanism to learn a nonlinear mapping from audio coefficients to lip landmark points. As shown in Fig.~\ref{fig deep VSG}(b), the model outputs the synthesized talking face video of former US President Barack Obama, following the pipeline of facial texture synthesis, video re-timing, and target video compositing. Beyond computer graphic video generation methods, as shown in Fig.~\ref{fig deep VSG}(c), Kumar~\etal~\cite{kumar2017obamanet} proposed the LSTM + UNet architecture, improving the model by replacing the complex video generation pipeline with a pix2pix framework~\cite{isola2017image}. In this way, there is no need to get involved with the details of a face,~\eg, synthesizing realistic teeth. However, as the above methods are trained on only Barack Obama with many hours of his weekly address footage, they cannot generalize to new identities or voices. The LSTM + UNet VSG backbone architecture is widely adopted in many subsequent works~\cite{sinha2020identity,wang2020mead,das2020speech,zhou2020makelttalk}. Unlike previous methods using audio MFCC features as input, Sinha~\etal~\cite{sinha2020identity,das2020speech} introduced DeepSpeech~\cite{hannun2014deep} features instead, as DeepSpeech features are more robust for the variation of speakers. 

In 2018, Jalalifar~\etal~\cite{jalalifar2018speech} proposed LSTM + C-GAN VSG backbone architecture, using the basic conditional generative adversarial network (C-GAN)~\cite{mirza2014conditional} for generating talking faces given the learned landmarks. As the LSTM network and C-GAN network are mutually independent, this model can reanimate the target face with audio from another person. In 2019, Chen~\etal~\cite{chen2019hierarchical} proposed a novel LSTM + Convolutional-RNN structure, further considering the correlation between adjacent video frames during generation. Besides, they also propose a novel dynamic pixel-wise Loss to solve the pixel jittering problem in correlated audio-visual regions. 
Wang~\etal~\cite{wang2021audio2head} proposed a three-stage VSG framework. Firstly, they use the 3D Hourglass Network as a motion field generator to predict landmark points based on the input audio, head motions, and the reference image. And then convert the predicted landmark points to dense motion fields. Finally, the synthesized talking video is obtained using a first-order motion model~\cite{siarohin2019first}. They recently updated the motion field generator by replacing the 3D Hourglass Network with a self-attention architecture~\cite{wang2021one}. 

Besides 2D landmark based approaches, mapping driving sources to 3D landmarks is also widely explored~\cite{zhou2020makelttalk,lu2021live,liu2023MODA}. Audio signals contain rich semantic-level information, including speech content, the speaker's speaking style, emotion,~\etc. Zhou~\etal~\cite{zhou2020makelttalk} utilized a voice conversion neural network to learn disentangled speech content and identity features. Then, an LSTM-based network is introduced to predict 3D landmarks based on speech content features. Finally, the final synthesized talking face video is realized using a UNet-style generator network. The key insight is to predict 3D landmarks from disentangled audio content features and speaker-aware features, such that they capture controllable lip synchronization and head motion dynamics. As shown in Fig.~\ref{fig deep VSG}(d), Lu~\etal~\cite{lu2021live} introduced extracting the high-level speech information using an autoregressive predictive coding (APC) model~\cite{chung2020generative} and manifold projection for better generalization. Then, an audio-to-lip-related motion module is designed to predict 3D lip landmarks. Finally, an image-to-image translation network (UNet) is introduced to synthesize video frames. 

\subsubsection{Coefficient based Methods} 
\textbf{2D Coefficient based.} Active Appearance Model (AAM) is one of the most commonly used facial coefficient models, representing both the shape and texture variations and their correlation. Fan~\etal~\cite{fan2015photo} utilized a two-layer BiLSTM network to estimate AAM coefficients of the mouth area based on the overlapped triphone input, which then is transferred to a face image to produce a photo-realistic talking head. The experiments show that the BiLSTM network has superior performance to previous HMM-based approaches. Similarly, as shown in Fig.~\ref{fig deep VSG}(e), Taylor~\etal~\cite{taylor2017deep} introduced a simple and effective DNN as a sliding window predictor to automatically learn AAM coefficients based on the fixed-length phoneme sequence. Furthermore, the model can be retargeted to drive other face models with the help of an effective retargeting approach. The main practical limitation of AAM coefficients is that the reference face AAM parameterization may cause potential errors when retargeting to a new subject. 

\textbf{3D Coefficient based.} Besides 2D facial coefficient models, 3D facial coefficients via principal component analysis (PCA) are more commonly used in VSG~\cite{pham2017speech,pham2018end,hussen2019speaker,thies2020neural,yi2020audio,lahiri2021lipsync3d,yao2021iterative}. Pham~\etal~\cite{pham2017speech,pham2018end,tzirakis2020synthesising} proposed utilizing CNN + RNN based backbone architectures to map audio signals to blendshape coefficients~\cite{cao2013facewarehouse} of a 3D face.
However, these methods rely heavily on the prior 3D facial models of target speakers.
Hussen~\etal~\cite{hussen2019speaker} finetuned a pretrained DNN-based acoustic model to map driving audios to 3D blendshape coefficients, as they hold the idea that a pretrained acoustic model has better generalization on speaker-independent VSG tasks than a randomly initialized model. As shown in Fig.~\ref{fig deep VSG}(f), Thies~\etal~\cite{thies2020neural} proposed a generalized Audio2Expression network and a specialized UNet-based neural face rendering network for audio-driven VSG. The proposed Audio2Expression network aims to estimate temporally stable 3D blendshape coefficients based on the DeepSpeech audio features, using a CNN based backbone architecture and a content-aware filtering network. In this way, the model is able to synthesize talking face videos from an audio sequence from another person.


Besides the 3D blendshape model, Kim~\etal~\cite{kim2018deep,kim2019neural} introduced 3D Morphable Model (3DMM)~\cite{deng2019accurate}, a more dense 3d face parametric representation, for video-oriented face2face translation. The 3DMM coefficients contain the rigid head pose parameters, facial identity coefficients, expression coefficients, gaze direction parameters for both eyes, and spherical harmonics illumination coefficients. Referring to the 3DMM based face2face translation pipeline mentioned above,~\cite{song2022everybody,yi2020audio,yao2021iterative,zhang2021flow,wu2021imitating,ji2021audio,zhang2021facial,zhang2023sadtalker} converted the driving source from video to audio clips (text scripts) and migrated this pipeline to VSG tasks. These methods have an approximate framework, as shown in Fig.~\ref{fig deep VSG}(g). The flowchart of this framework generally follows four steps: 1) Train a network to map the driving source to the facial expression coefficients, as visual speech information is implicit in facial expression coefficients. 2) Use a pretrained deep face reconstruction model to get the 3DMM coefficients of the reference identity image. 3) Combine the 3DMM coefficients from the reference identity image and the predicted facial expression coefficients to get hybrid 3DMM coefficients. 4) Synthesize talking videos using GPU rendering or a generation network.

Following the above flowchart, Song~\etal~\cite{song2022everybody} designed a novel Audio2Expression network. They empirically find that source identity information embedded in speech features will degrade the performance of mapping speech to mouth movement. Therefore, they explicitly add an ID-Removing sub-network to remove the identity information from the driving audio. Meanwhile, a UNet-style generation network is introduced to complete the mouth region guided by mouth landmarks. Yi~\etal~\cite{yi2020audio} proposed an LSTM-based network to map audio MFCC features to facial expression and head pose, as they argue that audio and head pose are correlated in a short period. Besides, they propose a memory-augmented GAN to refine these synthesized video frames into real ones. Wu~\etal~\cite{wu2021imitating} proposed an arbitrary talking style imitation VSG method. During the mapping stage, they introduced an extra style reference video as input and used a deep 3D reconstruction model to get the style code of the reference video. Next, they concatenate audio features with the reconstructed style code to predict the stylized 3DMM coefficients. However, the above 3DMM based models are not able to disentangle visual speech information from other facial expressions like eyebrow and head pose. Therefore, Zhang~\etal~\cite{zhang2021flow} proposed a novel flow-guided VSG framework, including one style-specific animation generator and one flow-guided video generator, to synthesize high visual quality videos. Moreover, the style-specific animation generator successfully disentangles lip dynamics with eyebrow and head pose. Li~\etal~\cite{li2021write} employed a similar framework for text-driven VSG. Ji~\etal~\cite{ji2021audio} proposed an emotional video portrait (EVP) to achieve audio-driven emotional control for talking face synthesis. Unlike previous methods, they adopt the cross-reconstruction~\cite{aberman2019learning} technique in the audio2expression stage to decompose the input audio into disentangled content and emotion embeddings. However, Mapping driven source to 3DMM coefficients directly suffers from the regression-to-mean problem. To overcome this issue, Xing~\etal~\cite{xing2023codetalker} introduced a pre-trained codebook to learn 3D motion priors instead of learning 3DMM coefficients.

\subsubsection{Vertex based Methods}
3D facial vertices are another popularly used 3D face model in VSG. For example, Karras~\etal~\cite{karras2017audio} used a simple CNN-based architecture to learn a nonlinear mapping from input audios to the 3D vertex coordinates (total 15,066 vertices) of the target face. To make the synthesized video more natural, they introduce an extra emotion code as an intuitive control for the emotional state of the face puppet. However, the proposed model is specialized for a particular speaker. To overcome this issue, as shown in Fig.~\ref{fig deep VSG}(h), Cudeiro~\etal~\cite{cudeiro2019capture} extended the model to multiple subjects. The proposed VOCA model concatenates the Deepspeech audio features and one-hot vector of a speaker and outputs 3D vertex (total 5023 vertices) displacements instead of vertex coordinates. The critical contribution of VOCA is that the additional identity control parameters can vary the identity-dependent visual dynamics. Based on VOCA, Liu~\etal~\cite{liu2021geometry} proposed a geometry-guided dense perspective network (GDPnet) with two constraints from different perspectives to achieve a more robust generation. Fan~\etal~\cite{fan2022faceformer} proposed a Transformer-based autoregressive VSG model named FaceFormer to encode the long-term audio context information and predict a sequence of 3D face vertices.

Richard~\etal~\cite{richard2021meshtalk} proposed a categorical latent space for VSG that disentangles audio-correlated and audio-uncorrelated (facial expressions like eye blinks, eyebrow) information based on a cross-modality loss. Then, a UNet-style architecture with skip connections is used to predict 3D vertex coordinates. Since the modalities disentanglement mechanism, the plausible motion of uncorrelated regions of the face is controllable, making the synthesized video more photo-realistic. Lahiri~\etal~\cite{lahiri2021lipsync3d} proposed a speaker-dependent VSR method, which decomposes the audio to talking face mapping problem into the prediction of the 3D face shape and the regressions over the 2D texture atlas. To do so, they first introduced a normalization preprocessing stage to eliminate the effects of head movement and lighting variations. Then, a geometry decoder and an auto-regressive texture synthesis network were trained to learn vertex displacements and the corresponding lip-centered texture, respectively. Finally, a computer graphics based video rendering pipeline is used to generate talking videos for the target speaker.

\subsection{One-Stage VSG Frameworks}
\label{subsec os vsg}
The two-stage VSG frameworks have been dominated before 2018. Nevertheless, two-stage VSG frameworks suffer from the complex processing pipeline, expensive and time-consuming facial parameter annotations, extra aid technologies like facial landmark detection and monocular 3D face reconstruction~\etc. Therefore, instead of optimizing the individual components of a complex two-stage VSG pipeline, researchers have paid more attention to exploring one-stage (end-to-end) VSG approaches. One-stage VSG pipelines refer to architectures that directly generate talking lip (face) videos from the driving source with an end-to-end learning strategy that does not involve any intermediate facial parameters. 

Speech2Vid~\cite{chung2017you} was among the first to explore one-stage VSG frameworks. As shown in Fig.~\ref{fig deep VSG}(i), it consists of four sub-networks. An audio encoder aims to extract speech features based on the driving audio; An identity encoder aims to extract identity features based on the reference image; And an image decoder tries to output synthesized images based on the fused speech and identity features. The above sub-networks form an autoencoder architecture, and L1 reconstruction loss is used for training. Besides, a separate pretrained deblurring CNN is introduced as a post-processing module to improve image quality. As a pioneer work, Speech2Vid provides a baseline for speaker-independent VSG and greatly motivates the research on one-stage VSG. 
However, Speech2Vid only uses the L1 reconstruction loss during training, which is inefficient for VSG for the following reasons. 1) The L1 reconstruction loss is operated on the whole face, and spontaneous motion of the face mainly occurs on the upper part of the face, leading to discouraging visual speech generation. 2) As Speech2Vid is temporal-independent (no knowledge of its previous outputs), it usually produces less coherent video sequences. 3) No consideration of consistency of the generated video with the driving audio. 

\subsubsection{GAN based Methods}
To overcome the above limitations of Speech2Vid, many researchers try to improve VSG performance by utilizing generative adversarial training~\cite{goodfellow2014generative} strategies. As shown in Fig.~\ref{fig deep VSG}(j), GAN based VSG methods usually consist of three sub-architectures,~\ie, encoders, generators, and discriminators.

Taking audio-driven VSG as an example, a piece of audio naturally entangles various information, such as speech, emotion, speaking style,~\etc. As we have emphasized in Section.~\ref{Generation-Related Challenges}, information coupling brings enormous challenges to VSR. To ameliorate this issue, Zhou~\etal~\cite{zhou2019talking} proposed a novel VSG framework called Disentangled Audio-Visual System (DAVS). Compared with previous VSG approaches, they focus more on the disentangled speech and identity feature extraction, which is realized based on supervised adversarial training. However, DAVS relies on extra Word-ID labels and Person-ID labels in the training stage. Sun~\etal~\cite{sun2021speech2talking} improved the model by learning speech and identity features within a self-supervised contrastive learning framework, with no need for extra annotations. Si~\etal~\cite{si2021speech2video} utilized knowledge distillation to disentangle emotion features, identity features, and speech features from the audio input with the help of a pretrained emotion recognition teacher network and a pretrained face recognition teacher network.
Recently, some works have tried to encode additional facial controllable dynamics like emotion and head poses into the generation pipeline to generate a more natural-spontaneous talking face. For example,~\cite{sadoughi2019speech, eskimez2021speech} introduce additional emotion encoders, and~\cite{zhou2021pose,liang2022expressive,wang2022one} devise additional pose encodings into the generation pipeline.

Considering the drawbacks of only using image reconstruction loss, GAN based methods focus on customizing more effective learning goals for VSG. For example, Prajwal~\etal~\cite{kr2019towards,prajwal2020lip} introduced a simple audio-visual synchronization discriminator for lip-syncing VSG. In addition, Chen~\etal~\cite{chen2018lip} proposed an audio-visual derivative correlation loss to optimize the consistency of the two modalities in feature spaces and a three-stream GAN discriminator to force talking mouth video generation depending on the input audio signal. 

For temporal-dependent video generation,~\cite{vougioukas2018end,vougioukas2020realistic,eskimez2020end} utilized autoregression style VSG generator networks for talking face generation. Two discriminators,~\ie, a frame and sequence discriminator, are used to optimize the generated facial dynamics. Based on~\cite{vougioukas2018end}, Song~\etal~\cite{song2019talking} introduced a VSR discriminator further to improve the lip movement accuracy of generated talking videos. The ablation study demonstrated that the additional VSR discriminator helps achieve more obvious lip movement, proving our motivation that VSR and VSG are dual and mutually promoted. Furthermore, Chen~\etal~\cite{chen2020duallip} developed the DualLip system to jointly improve VSR and VSG by leveraging the task duality and demonstrating that both VSR and VSG models can be enhanced with the help of extra unlabeled data.
Besides the above learning goals, the optical flow discriminator~\cite{zeng2020talking}, speech-related facial action units~\cite{chen2021talking}, and cross-modal mutual information estimator~\cite{zhu2021arbitrary} are also utilized to optimize lip motion and cross-modal consistency of generated talking videos with the driving source. 

\begin{table*}
    \centering
    \caption{The comparison of representative VSG methods.}
    \label{VSG comparison}
    \resizebox{\textwidth}{!}{%
    \begin{tabular}{|c|c|c|c|c|c|c|c|c|c|c|c|c|c|} 
    \hline
    \multicolumn{3}{|c|}{\multirow{2}{*}{Framework \& Method}}                                                                                                                                        & \multicolumn{4}{c|}{Settings}                                                                                                                                              & \multicolumn{3}{c|}{GRID} & \multicolumn{3}{c|}{LRW} & \multirow{2}{*}{Highlights}                                                                                                                                     \\ 
    \cline{4-13}
    \multicolumn{3}{|c|}{}                                                                                                                                                                          & Input$^{\dagger}$ & Training Set                                                     & Extra Requriment                                                                 & By-product$^{\ddagger}$    & PSNR  & SSIM & LMD        & PSNR   & SSIM  & LMD     &                                                                                                                                                                 \\ 
    \hline
    \multirow{11}{*}{\begin{tabular}[c]{@{}c@{}}\rotatebox{90}{Two-stage} \end{tabular}} & \multirow{7}{*}{\begin{tabular}[c]{@{}c@{}}Landmark \\based\end{tabular}}    & Chen~\etal~\cite{chen2019hierarchical}                & A+I   & \begin{tabular}[c]{@{}c@{}}LRW \\GRID\end{tabular}               & Landmark detector                                                              & /            & 32.15 & 0.83 & 1.29       & 30.91  & 0.81  & 1.37    & \begin{tabular}[c]{@{}c@{}}Proposed a Convolutional-RNN structure, \\which utilizes correlation between adjacent \\frames in the generation stage\end{tabular}  \\ 
    \cline{3-14} &                                                                              & Das~\etal~\cite{das2020speech}             & A+I   & TCD-TIMIT                                                        & \begin{tabular}[c]{@{}c@{}}Landmark detector;\\DeepSpeech model\end{tabular}     & Blink motion & 29.9  & 0.83 & 1.22       & /      & /     & /       & \begin{tabular}[c]{@{}c@{}}Proposed two GAN based networks to \\ learn the motion and texture separately\end{tabular}                                  \\ 
    \cline{3-14}
    &                                                                              & Wang~\etal~\cite{wang2021audio2head}           & A+I   & \begin{tabular}[c]{@{}c@{}}GRID\\LRW\\VoxCeleb\end{tabular}      & \begin{tabular}[c]{@{}c@{}}Landmark detector;\\Pretrained image generator\\ and face encoder\end{tabular} & Head Motion  & 30.93 & 0.91 & /          & 19.53* & 0.63* & /       & \begin{tabular}[c]{@{}c@{}}Regressed the head motions \\ in accordance with audio dynamics\end{tabular}                                              \\ 
    \cline{2-14}
    & \multirow{4}{*}{\begin{tabular}[c]{@{}c@{}}Coefficient \\based\end{tabular}} & Song~\etal~\cite{song2022everybody} & A+V   & \begin{tabular}[c]{@{}c@{}}GRID \& \\a novel dataset\end{tabular}    & Face reconstruction                                                              & Head Motion  & 32.23 & 0.97 & /          & /      & /     & /       & \begin{tabular}[c]{@{}c@{}}Proposed an Audio ID-Removing Network \\for pure speech feature learning\end{tabular}                                                 \\ 
    \cline{3-14}
    &                                                                              & Yi~\etal~\cite{yi2020audio}             & A+I   & LRW                                                              & Face reconstruction                                                              & Head Motion  & /     & /    & /          & 30.94  & 0.75  & 1.58    & \begin{tabular}[c]{@{}c@{}}~ ~Proposed a memory-augmented GAN \\ module for rendered frames refining\end{tabular}                                                \\ 
    \hline
    \multirow{19}{*}{\begin{tabular}[c]{@{}c@{}}\rotatebox{90}{One-stage}\end{tabular}} & \begin{tabular}[c]{@{}c@{}}Auto\\Encoder\end{tabular}                        & Chung~\etal~\cite{chung2017you}        & A+I   & \begin{tabular}[c]{@{}c@{}}VoxCeleb\\LRW\end{tabular}            & Pretrained face encoder                                                          & /            & 29.36 & 0.74 & 1.35       & 28.06  & 0.46  & 2.25    & promoted the research on one-stage VSG                                                                                                                \\ 
    \cline{2-14}
    & \multirow{15}{*}{\begin{tabular}[c]{@{}c@{}}GAN \\Based\end{tabular}}         & Vougioukas~\etal~\cite{vougioukas2020realistic}     & A+I   & GRID                                                             & Pretrained VSR model                                                                               & /            & 27.10 & 0.82 & /          & 23.08  & 0.76  & /       & \begin{tabular}[c]{@{}c@{}} Utilized an autoregressive temporal GAN \\ for more coherent sequences generation \\ \end{tabular}                                                  \\ 
    \cline{3-14}
    &                                                                              & Chen~\etal~\cite{chen2018lip}    & A+I   & \begin{tabular}[c]{@{}c@{}}GRID\\LRW\end{tabular}                & Pretrained FlowNet                                                               & /            & 29.89 & 0.73 & 1.18       & 28.65  & 0.53  & 1.92    & \begin{tabular}[c]{@{}c@{}}Proposed a novel correlation loss to \\synchronize lip movements and input audio\end{tabular}                                         \\ 
    \cline{3-14}
    &                                                                              & Prajwal~\etal~\cite{kr2019towards}     & A+V   & LRS2                                                             & /                                                                                & /            & /     & /    & /          & 33.4   & 0.96  & 0.60    & \begin{tabular}[c]{@{}c@{}}Proved that a lip synchronization \\ discriminator is quite useful for VSG\end{tabular}                                                     \\ 
    \cline{3-14}
    &                                                                              & Song~\etal~\cite{song2019talking}      & A+I   & \begin{tabular}[c]{@{}c@{}}TCD-TIMIT\\LRW\\VoxCeleb\end{tabular} & Pretrained VSR model                                                             & /            & /     & /    & /          & 27.43  & 0.92  & 3.14    & \begin{tabular}[c]{@{}c@{}}Introduced a lip-reading discriminator \\to guide lip motion generation\end{tabular}                                   \\ 
    \cline{3-14}
    &                                                                              & Zhou~\etal~\cite{zhou2019talking}           & A+V   & LRW                                                              & Word and Identity labels                                                         & /            & /     & /    & /          & 26.7   & 0.88  & /       & \begin{tabular}[c]{@{}c@{}}Improved visual quality by using disentangled\\~audio-visual representation learning\end{tabular}                                          \\ 
    \cline{3-14}
    &                                                                              & Zhu~\etal~\cite{zhu2021arbitrary}       & A+I   & \begin{tabular}[c]{@{}c@{}}GRID\\LRW\end{tabular}                & /                                                                                & /            & 31.01 & 0.97 & 0.78       & 32.08  & 0.92  & 1.21    & \begin{tabular}[c]{@{}c@{}}Improved cross-modality coherence with a novel \\ Asymmetric Mutual Information Estimator (AMIE)\end{tabular}                               \\ 
    \cline{3-14}
    &                                                                              & Chen~\etal~\cite{chen2021talking}            & A+I   & \begin{tabular}[c]{@{}c@{}}GRID\\TCD-TIMIT\end{tabular}          & AU Classifier                                                                    & /            & 29.84 & 0.77 & /          & /      & /     & /       & \begin{tabular}[c]{@{}c@{}}Used both audio and speech-related facial \\action units (AUs) as driving information\end{tabular}                                    \\ 
    \cline{2-14}
    & Others                                                                       & Ye~\etal~\cite{ye2022audio}            & A+V   & a mixed dataset                                                  & Pretrained AudioNet                                                              & /            & /     & /    & /          & 31.98  & 0.81  & 1.44    & \begin{tabular}[c]{@{}c@{}}Proposed a novel one-stage VSG paradigm with \\the introduction of dynamic convolution kernels\end{tabular}                            \\
    \hline
    \end{tabular}
    }
    \begin{tablenotes}
    \item[1] $*$: Including background regions. 
    \item[2] $^{\dagger}$: A-Audio, I-Image, V-Video.
    \item[3] $^{\ddagger}$: Additional effects besides VSG.
    \end{tablenotes}
    \end{table*}

\subsubsection{Other Methods}
In addition, some other one-stage VSG schemes have also been proposed. Inspired by the success of the neural radiance field (NeRF)~\cite{mildenhall2020nerf}, Guo~\etal~\cite{guo2021ad} proposed the audio-driven neural radiance fields (AD-NeRF) model for VSG. As shown in Fig.~\ref{fig deep VSG}(k), AD-NeRF takes DeepSpeech audio features as conditional input, learning an implicit neural scene representation function to map audio features to dynamic neural radiance fields for talking face rendering. Furthermore, AD-NeRF models not only the head region but also the upper body via learning two individual neural radiance fields. However, AD-NeRF does not generalize well on mismatched driving audios and speakers. Motivated by the excellent 3D face structural information modeling ability, more and more works~\cite{shen2022learning,tang2022real,liu2022semantic, ye2022geneface} tried to utilize NeRF technology for VSG. As shown in Fig.~\ref{fig deep VSG}(l), unlike the previous concatenation-based feature fusion strategy, Ye~\etal~\cite{ye2022audio} presented a fully convolutional neural network with dynamic convolution kernels (DCKs) for cross-modal feature fusion, which extracts features from audio and reshapes features as DCKs of the fully convolutional network. Due to the simple yet effective network architecture, the real-time performance of VSG is significantly improved.

\subsection{Summary and Performance Comparison}
\label{VSG Summary}
Visual speech generation is an important and challenging problem in the cross-field of computer vision, computer graphics, and natural language analysis and has received considerable attention in recent five years. Moreover, thanks to remarkable developments in deep learning techniques, the field of VSG has dramatically evolved. In this subsection, we will discuss representative VSG methods on large-scale datasets and summarize the main issues of VSG.

Because VSG approaches have various implementation requirements (driving sources, extra technologies, diverse annotation needs, specific datasets,~\etc.) and configurations (training sets, learning paradigms, lip or whole face generation, background, pose and emotion control,~\etc.), it may be impractical to compare every recently proposed VSG method in a unified and fair manner.

It is nevertheless valuable to integrate some representative VSG methods and their requirements, configurations, and highlights into a table. Therefore, as shown in Table.~\ref{VSG comparison}, we summarize the performance and experimental settings of some representative VSG methods tested on large-scale, commonly used benchmark datasets, including GRID and LRW. 

To give readers a general understanding of the performance of the VSG method in different frameworks, three commonly used quantitative evaluation metrics,~\ie, PSNR, SSIM, and LMD, are listed in Table.~\ref{VSG comparison}. It is worth noting that the above three metrics are most widely used for VSG, even though they are not yet effective and perfect enough. Although many quantitative metrics for VSG were proposed recently, the following issues need further investigation.
\begin{itemize}
\renewcommand{\labelitemi}{$\bullet$}
    \item In the early stage of VSG, qualitative evaluations are primarily utilized, such as visualization and user preference studies~\cite{taylor2017deep,garrido2015vdub,suwajanakorn2017synthesizing}. However, qualitative evaluations are unstable and unreproducible. 
    \item Many works have attempted to establish VSG evaluation benchmarks, and more than a dozen evaluation metrics have been proposed for VSG. Consequently, existing VSG evaluation benchmarks are not unified. Chen~\etal~\cite{chen2020comprises} have conducted a survey of VSG evaluation and designed a unified benchmark for VSG according to desired properties. To promote VSG development, researchers should pay more effort to VSG evaluation benchmarks.
    \item The results of quantitative evaluation and qualitative evaluation are sometimes in mutual conflict. For example, some works~\cite{song2019talking,vougioukas2020realistic,wang2021audio2head} have observed that both the PSNR and SSIM are negatively affected by introducing image or video discriminators. Nevertheless, these discriminators significantly improve the video realism and visual quality in the user study experiments. 
    \item In practical applications, real-time is another substantial demand for VSG. However, most of the current VSG methods ignore real-time. Therefore, real-time performance is also an important evaluation metric that needs to be considered in the future.
\end{itemize}

\section{Conclusion and Outlooks}
\label{sec conclusion}
In this paper, we have presented a comprehensive review of the deep learning based VSA. We focus on two fundamental questions,~\ie, visual speech recognition and visual speech generation, and summarize realistic challenges and current developments, including datasets, evaluation protocols, representative methods, SOTA performance, practical issues,~\etc. We presented a systemic overview of VSR and VSG approaches and discussed their underlying connections, contributions, and shortcomings. Considering that many practical issues discussed in Section.~\ref{VSR Summary} and Section.~\ref{VSG Summary} remain unresolved, there are still enough opportunities for VSA research and application. We attempt to provide some ideas and discuss potential future research directions in the following.


\textbf{Learning with fewer labels.} As we have mentioned above, collecting a large-scale audio-visual dataset is quite costly, and manually labeling is even more time-consuming. Existing deep learning based VSA approaches usually rely heavily on labeled data, which is a current limitation of VSA research. Recently, some works have explored cross-modal self-supervised learning, knowledge distillation for VSA. However, it is valuable to explore other label-efficient learning paradigms like domain adaptation, active learning, few-shot learning,~\etc.

\textbf{Multilingual VSA.} Existing audio-visual datasets are mostly monolingual. In general, English is the most universal language. However, in some practical scenes like air traffic control (ATC) and international conferences, multilingual communication is needed. Although multilingual audio speech recognition has been widely explored, multilingual visual speech recognition has received little attention yet.

\textbf{Extended applications of VSA.} Besides VSR and VSG, there are also some hot topics that VSA can be helpful. One of the most common tasks is audio-visual speech recognition (AVSR), a speech recognition technology that uses visual and audio information. Another typical extended task is audio-visual speech enhancement (AVSE), aiming to separate a speaker's voice given lip regions in the corresponding video by predicting both the magnitude and the phase of the target signal. Besides, for DeepFake detection, VSA can serve as an effective technology for counterfeit talking video detection.

\textbf{VSA technologies for virtual characters.} As an emerging type of internet application and social platform, Metaverse has recently gained a lot of attention. Virtual avatar modeling is a crucial technology in the field of Metaverse. With the rapid development of Metaverse technology, virtual characters-oriented VSA technologies also came into being. Considering that existing VSA methods mostly focus on realistic speakers, virtual characters-oriented VSA research is a potential direction in the future.

\textbf{Security \& robustness for VSA.} Security and robustness are important requirements in the public safety field of VSA technology. Recent research has demonstrated that deep learning-based AI systems are vulnerable to different types of attacks, such as
adversarial attacks~, and spoofing attacks. This raises serious concerns in the field of security. However, security and robustness are not taken seriously in existing VSA approaches. 

\textbf{Privacy preserving for VSA.} As VSA involves face-related private information, it is hard to construct a public large-scale audio-visual dataset, which also hinders the development of VSA. 
To address this issue, available privacy-preserving techniques such as Federated Learning, Homomorphic Encryption, and Secure Multi-Party Computation can be helpful. However, to the best of our knowledge, privacy-preserving VSA research has not yet started.

\footnotesize
\bibliographystyle{IEEEtran}
\bibliography{IEEEabrv,lilibib}

\newpage

\end{document}